\def\paperID{5056}
\def\confName{CVPR}
\def\confYear{2024}

\def\paperTitle{\method: Probabilistic Photorealistic 3D Reconstruction of Humans}

\def\authorBlock{
    Akash Sengupta$^{1,2}$\thanks{Work done as an intern at Google Research.} \qquad Thiemo Alldieck$^1$ \qquad Nikos Kolotouros$^1$ \qquad Enric Corona$^1$ \\ Andrei Zanfir$^1$ \qquad Cristian Sminchisescu$^1$
}

\newif\ifreview 
\newif\ifarxiv \newcommand{\arxiv}{\arxivtrue}
\newif\ifcamera 
\newif\ifrebuttal 

\arxiv %

\pdfoutput=1
\documentclass[10pt,twocolumn,letterpaper]{article}
\ifreview \usepackage[review]{cvpr} \fi
\ifarxiv \usepackage[pagenumbers]{cvpr} \fi
\ifrebuttal \usepackage[rebuttal]{cvpr} \fi
\ifcamera \usepackage{cvpr} \fi

\usepackage{graphicx}	
\usepackage{amsmath}	
\usepackage{amssymb}	
\usepackage{booktabs}
\usepackage{times}
\usepackage{microtype}
\usepackage{epsfig}
\usepackage[table,xcdraw,dvipsnames]{xcolor}
\usepackage[font=small,skip=5pt,belowskip=-5pt]{caption}
\usepackage{float}
\usepackage{placeins}
\usepackage{color, colortbl}
\usepackage{stfloats}
\usepackage{enumitem}
\usepackage{tabularx}
\usepackage{xstring}
\usepackage{multirow}
\usepackage{xspace}
\usepackage{url}
\usepackage{subcaption}
\usepackage[hang,flushmargin]{footmisc}

\definecolor{limegreen}{HTML}{badc58}
\definecolor{myyellow}{HTML}{f6e58d}
\definecolor{babyblue}{rgb}{0.54, 0.81, 0.94}

\newcommand{\cbest}[1]{\cellcolor{limegreen}\textbf{#1}}
\newcommand{\csecond}[1]{\cellcolor{myyellow}#1}
\newcommand{\best}[1]{\colorbox{limegreen}{\textbf{#1}}}
\newcommand{\second}[1]{\colorbox{myyellow}{#1}}

\ifcamera \usepackage[accsupp]{axessibility} \fi

\newcommand{\nbf}[1]{{\noindent \textbf{#1.}}}
\renewcommand{\paragraph}[1]{\vspace{0.5mm}\noindent\textbf{#1}\:}

\newcommand{\method}{DiffHuman\xspace}
\newcommand{\supp}{Suppl.\ Mat\onedot}
\ifarxiv \renewcommand{\supp}{appendix\xspace} \fi

\newcommand{\R}[1]{{%
    \textbf{%
        \ifstrequal{#1}{1}{\textcolor{red}{R#1}}{%
        \ifstrequal{#1}{2}{\textcolor{blue}{R#1}}{%
        \ifstrequal{#1}{3}{\textcolor{magenta}{R#1}}{%
        \ifstrequal{#1}{4}{\textcolor{teal}{R#1}}{%
                           \textcolor{cyan}{R#1}%
        }}}}%
    }%
}}

\usepackage{xr-hyper}

\makeatletter
\newcommand*{\addFileDependency}[1]{
  \typeout{(#1)}
  \@addtofilelist{#1}
  \IfFileExists{#1}{}{\typeout{No file #1.}}
}

\makeatother

\definecolor{cvprblue}{rgb}{0.21,0.49,0.74}
\usepackage[pagebackref,breaklinks,colorlinks,citecolor=cvprblue]{hyperref}
\usepackage[capitalize]{cleveref}
\crefname{section}{Sec.}{Secs.}
\crefname{table}{Table}{Tables}
\crefname{figure}{Fig.}{Figs.}

\begin{document}
\setlength{\abovedisplayskip}{3pt}
\setlength{\belowdisplayskip}{3pt}

\title{\paperTitle\vspace{-3mm}}
\author{\authorBlock}

\makeatletter
\let\@oldmaketitle\@maketitle%
\renewcommand{\@maketitle}{
    \@oldmaketitle%
    \centering
    \iftoggle{cvprfinal}{
        \centering
        \vspace{-8mm}
        \normalsize{$^1$Google Research \quad $^2$University of Cambridge}
        \vspace{4mm}
        \\
    }{\vspace{-10mm}}
    \centering
    \centering
\includegraphics[width=0.96\linewidth,trim={0 5.6cm 0 1.5cm},clip]{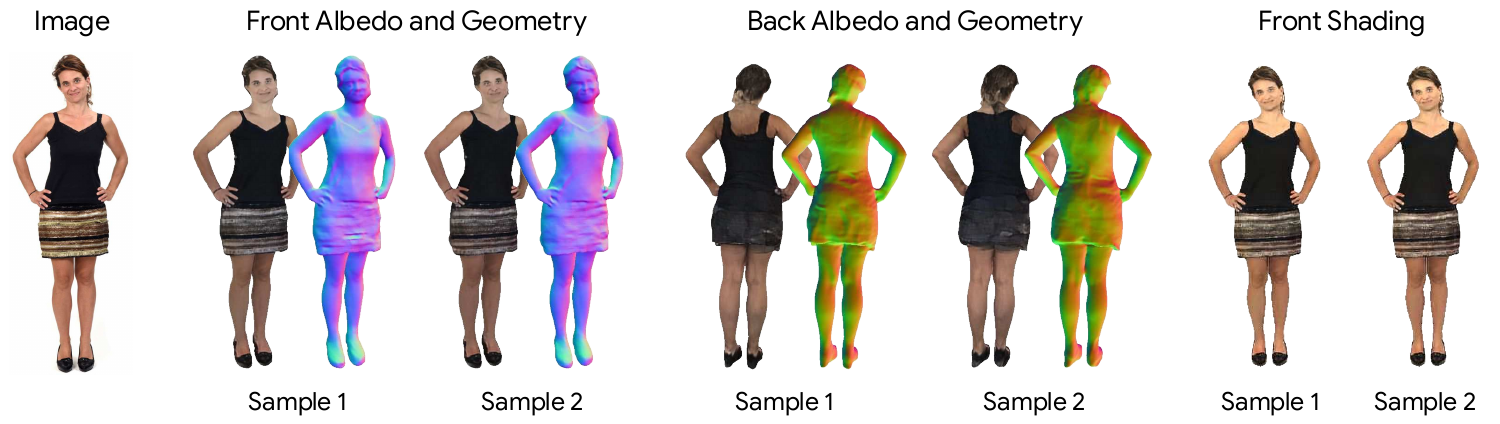}
\captionof{figure}{\method predicts a probability distribution over 3D human reconstructions conditioned on a single monocular RGB image. This enables us to sample multiple plausible, diverse and input-consistent reconstructions during inference. Samples from \method demonstrate a high level of geometric and colour-wise detail, particularly in unseen and uncertain regions of the human body surface.}
\label{fig:teaser}

    \vspace{6mm}
}
\makeatother

\maketitle
\begin{abstract}
\vspace{-3mm}
We present \method, a probabilistic method for photorealistic 3D human reconstruction from a single RGB image. 
Despite the ill-posed nature of this problem, most methods are deterministic and output a single solution, often resulting in a lack of geometric detail and blurriness in unseen or uncertain regions. 
In contrast, \method predicts a \textbf{probability distribution} over 3D reconstructions conditioned on an input 2D image, which allows us to sample multiple detailed 3D avatars that are consistent with the image.
\method is implemented as a conditional diffusion model that denoises pixel-aligned 2D observations of an underlying 3D shape representation.
During inference, we may sample 3D avatars by iteratively denoising 2D renders of the predicted 3D representation. 
Furthermore, we introduce a generator neural network that approximates rendering with considerably reduced runtime ($55\times$ speed up), resulting in a novel dual-branch diffusion framework.
Our experiments show that \method can produce diverse and detailed reconstructions for the parts of the person that are unseen or uncertain in the input image, while remaining competitive with the state-of-the-art when reconstructing visible surfaces.

\end{abstract}
\vspace{-4mm}

\section{Introduction}
\label{sec:intro}

Photorealistic 3D reconstruction of humans from a single image is a central problem for a wide range of applications.
Avatar creation for virtual and mixed reality, games, movie production, or fitness and health applications all benefit from reliable and easy-to-use 3D human reconstruction.
However, monocular 3D reconstruction is ill-posed:
depth-ambiguities, (self)-occlusion, and unobserved body parts make it infeasible to reconstruct the true, veridical 3D shape and appearance of the subject.
In fact, there exist an infinite number of 3D scenes that could have produced a given image; although not all of them would represent plausible human and clothing geometry, and realistic interplay between physical albedo and lighting.
Yet, existing methods \cite{saito2020pifuhd,xiu2022icon,alldieck2022phorhum,corona2023s3f} still treat the problem as a one-to-one mapping and return just \emph{one} plausible 3D solution.
Simply assuming that this single solution is correct can lead to failures in downstream usages of the 3D reconstruction.
Moreover, deterministic methods often produce detail-less reconstructions of unobserved or uncertain surface regions, \eg the back of a person.
This is a well-known effect of applying deterministic training losses to ill-posed learning problems \cite{bishop94mixturedensity, cui2020learning, mathieu2015deep}, which causes predictions to fall back towards the mean of the underlying training distribution when faced with ambiguity. The mean may not have high probability in a multi-modal distribution and often represents blurry and over-smooth 3D reconstructions.

In this work, we overcome the shortcomings of deterministic methods by predicting a \emph{distribution} over possible 3D human reconstructions.
Our method \textbf{\method} uses a single input image to condition a denoising diffusion model \cite{ho2020denoising}, which generates pixel-aligned front and back observations of the underlying 3D human.
To enable full 3D reconstruction, we take inspiration from recent work \cite{tewari2023diffusion, szymanowicz23viewset_diffusion} and integrate rendering of an intermediate implicit 3D representation into the model's denoising step.
This allows us, at test time, to reconstruct 3D meshes from a signed distance and colour field defined by this same intermediate representation.
However, diffusion-via-rendering is notoriously slow. Thus, we develop a hybrid solution which replaces the expensive implicit surface rendering with a single forward pass through an additional generator network, resulting in a $55\times$ speed up at test time.
Our probabilistic approach enables us to sample multiple input-consistent reconstructions and visualise prediction uncertainty, while significantly improving the quality of unseen surfaces.
In summary, our contributions are:
\begin{itemize}
    \item[-] We present a probabilistic diffusion model for photorealistic 3D human reconstruction that predicts a distribution of plausible reconstructions conditioned on an input image.
    \item[-] We propose a novel dual-branch framework that utilises an image generation network, alleviating the need for expensive implicit surface rendering at every denoising step.
    \item[-] We show that our model produces 3D reconstructions with greater levels of geometric detail and colour sharpness in uncertain regions than the current state-of-the-art.
\end{itemize}

\begin{figure*}[tp]
    \vspace{-4.5mm}
    \centering
    \includegraphics[width=\linewidth,trim={0 9.2cm 0 0},clip]{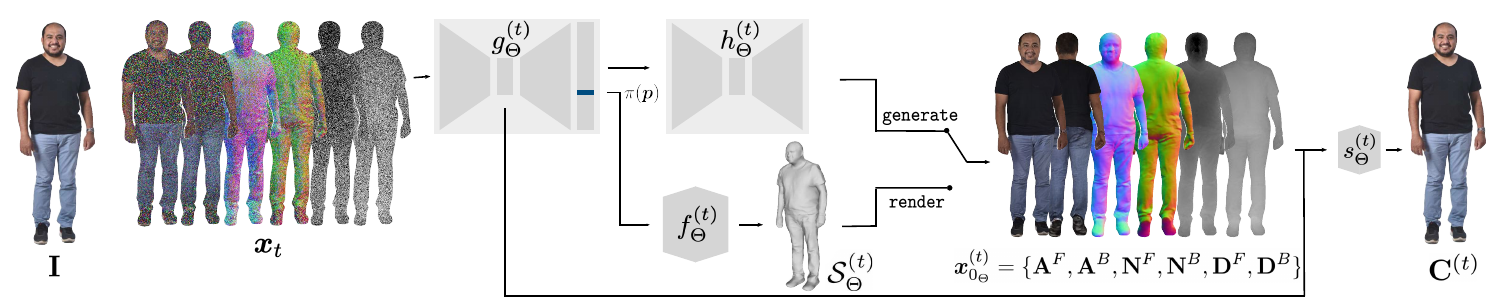}
    \caption{
    \textbf{Method overview}. We use a diffusion probabilistic model \cite{ho2020denoising} to predict a distribution over plausible 3D reconstructions conditioned on a single RGB image. During training, we predict noise-dependent pixel-aligned features $g^{(t)}_\Theta(\boldsymbol x_t, \mathbf I)$ given a noisy observation set $\boldsymbol x_t$ consisting of front/back albedo, depth and normal renders, and an RGB image $\mathbf{I}$. These features condition an SDF $f^{(t)}_\Theta$, which is dependent on both $\boldsymbol x_t$ and $\mathbf{I}$. $f^{(t)}_\Theta$ and $g^{(t)}_\Theta$ are neural networks that define an implicit surface $\mathcal{S}^{(t)}_\Theta(\mathbf{\boldsymbol x_t, I})$. Then, we obtain an estimate of the denoised observation set $\boldsymbol x_{0_\Theta}^{(t)}$ by rendering $\mathcal{S}^{(t)}_\Theta$. We may additionally produce a shaded image $\mathbf{C}^{(t)}$ by applying a pixel-wise noise-dependent shading network $s^{(t)}_\Theta$.
    During inference, we can sample trajectories over observation sets $\boldsymbol x_{0:T} \sim p_\Theta(\boldsymbol x_{0:T} | \mathbf{I})$ by computing and rendering $\mathcal{S}^{(t)}_\Theta(\mathbf{\boldsymbol x_t, I})$ in each denoising step. Our final 3D samples $\mathcal{S} \sim p_\Theta(\mathcal{S}| \mathbf{I})$ are obtained as the final reconstruction $\mathcal{S} = \mathcal{S}_\Theta^{(1)}(\boldsymbol x_1, \mathbf I)$. To mitigate the computational cost of rendering an implicit surface in every step, we train a ``generator'' network $h_\Theta^{(t)}$ that imitates rendering by directly mapping $g^{(t)}_\Theta(\boldsymbol x_t, \mathbf I)$ to $\boldsymbol x_{0_\Theta}^{(t)}$. During inference, we denoise using $h_\Theta^{(t)}$ and only explicitly compute the 3D reconstruction in the last step.
    }
    \label{fig:method}
    \vspace{-2mm}
\end{figure*}

\section{Related Work}
\label{sec:related}
We give an overview of related work on photorealistic and probabilistic 3D human reconstruction.

\paragraph{Photorealistic 3D human reconstruction.}
Methods for human reconstruction in 3D can be broadly categorised into three classes: mesh-based, implicit, and NeRF-based.

Several methods \cite{bogo2016smplify, tan2017, kanazawa2018end, pavlakos2018learning, omran2018neural, kolotouros2019learning, guler2019holopose, zhang2019danet, sengupta2020synthetic, kocabas2021pare, zanfir2021neural, zanfir2021thundr} attempt to reconstruct humans in 3D by leveraging parametric body models \cite{anguelov2005scape, smpl, xu2020ghum}.
However these approaches only reconstruct the body geometry under clothing and do not predict texture. Others focus on learning deformations on top of parametric models to model hair and clothing \cite{alldieck2018detailed, alldieck2018video, alldieck2019learning, alldieck2019tex2shape, lazova2019360, xiang2020monoclothcap, zhu2019detailed}. These methods are generally fast to render, but have the drawback of working with low resolution meshes that cannot capture fine geometric and texture details, and more importantly cannot handle garments with topologies that deviate from the body, such as skirts or dresses.

Implicit methods model 3D surface geometry as the level-set of a signed-distance function \cite{park2019deepsdf} or occupancy field \cite{mescheder2019occupancy}.
They are able to model surfaces of arbitrary topology, which makes them suitable for representing clothed humans.
PIFu \cite{saito2019pifu} and PIFuHD \cite{saito2020pifuhd} predict occupancy and colour fields directly from an input image using pixel-aligned features.
Geo-PIFu \cite{he2020geo} and PaMIR \cite{zheng2021pamir} use a combination of pixel-aligned features and sampled features from a voxel grid to mitigate depth ambiguity issues.
PHORHUM \cite{alldieck2022phorhum} replaces the occupancy field with a signed distance function and decouples albedo and shading. ARCH \cite{huang2020arch}, ARCH++ \cite{he2021arch++}, and S3F \cite{corona2023s3f} leverage a human body prior and reconstruct animatable avatars.
ICON \cite{xiu2022icon} only predicts surface geometry, using a normal refinement procedure that alternates between normal prediction and body pose refinement.
ECON \cite{xiu2023econ} independently reconstructs front and back surfaces, which are then fused using a body model prior.
TECH \cite{huang2024tech} is a concurrent optimisation-based method that uses guidance from a text-to-image diffusion model to reconstruct invisible surfaces.
DiffuStereo \cite{shao2022diffustereo} reconstructs detailed 3D human geometry using a multi-view stereo setup, whereas POSEFusion \cite{li2021posefusion} uses a single RGB-D camera. D-IF \cite{yang2023d} models uncertainty in occupancy field predictions based on the distance of a point from the surface.
In contrast, our method learns a distribution over plausible implicit surfaces for a given image from which we can sample at test time.

Another line of work for photorealistic 3D human reconstruction uses Neural Radiance Fields (NeRFs) \cite{mildenhall2021nerf} as the underlying representation. However, these often require multi-view setups or long videos to train \cite{xu2021h, jiang2022neuman, li2023posevocab, weng2022humannerf, isik2023humanrf}. Towards the task of monocular reconstruction, SHERF \cite{hu2023sherf} and ELICIT \cite{huang2023one} learn animatable NeRFs from a single image. They are both driven by an underlying body model.

\paragraph{Probabilistic 3D human reconstruction.}
Several methods estimate distributions over 3D poses conditioned on an input image.
For example, \cite{li2019generating} uses mixture density networks to estimate a distribution over 3D keypoints conditioned on observed 2D keypoint locations. \cite{wehrbein2021probabilistic} replaces mixture density networks with normalising flows.
More recent methods, such as \cite{gong2023diffpose} and \cite{shan2023diffusion}, employ diffusion models for learning a distribution over 3D poses.
3D Multibodies \cite{biggs20203d} predicts a categorical distribution over SMPL \cite{smpl} parameter hypotheses conditioned on an input image, while ProHMR \cite{kolotouros2021probabilistic} utilises conditional normalising flows to this end.
Sengupta \etal \cite{sengupta2021hierarchical} output hierarchical matrix-Fisher distributions that exploit the SMPL kinematic tree.
HuManiFlow \cite{sengupta2023humaniflow} predicts normalising flow distributions over ancestor-conditioned joint rotations, which respect the structure of the 3D rotation group $SO(3)$.
All these methods, however, predict distributions over sparse 3D landmarks, joint rotations or body model parameters. In contrast, our method learns a much more expressive distribution over detailed surfaces corresponding to clothed human geometry and appearance.

\paragraph{3D diffusion models.}
The success of diffusion models for 2D image synthesis \cite{saharia2022photorealistic, rombach2022high} has motivated a few methods that apply these to 3D generation.
A pertinent challenge in this task is the choice of an appropriate 3D representation.
\cite{luo2021diffusion, zeng2022lion} implement diffusion models for 3D point cloud generation.
DiffRF \cite{muller2023diffrf} learns a diffusion model for generating volumetric radiance fields, but denoising 3D voxel grids is computationally expensive. 
HyperDiffusion \cite{erkoc2023hyperdiffusion} presents a method for 3D shape generation that performs diffusion in the weight space of occupancy networks.
However, this requires offline fitting of an occupancy field to every training example.
\cite{tewari2023diffusion, szymanowicz23viewset_diffusion} integrate rendering of an intermediate 3D representation into the denoising step of a 2D diffusion model, which enables 3D sampling during inference. 
Our method is similar, but we mitigate the cost of diffusion-via-rendering using a 2D generator neural network.

\vspace{-1mm}
\section{Method}
\label{sec:method}
\vspace{-1mm}
This section details our method for predicting diffusion-based distributions over implicit surfaces representing human geometry and appearance.
We begin with an overview of denoising diffusion models and implicit surfaces.

\subsection{Background}
\vspace{-1mm}
\label{subsec:method_background_ddpm}
\paragraph{Denoising Diffusion Probabilistic Models.}
 (DDPMs) \cite{ho2020denoising} are generative models that learn to sample from a target data distribution $q(\boldsymbol x_0)$ via a learned iterative denoising process. A forward Markov chain $q(\boldsymbol x_{0:T})$ progressively adds Gaussian noise to data samples $\boldsymbol x_0 \sim q(\boldsymbol x_0)$ such that
\begin{equation}
    q(\boldsymbol x_t | \boldsymbol x_{t-1}) = \mathcal{N}(\boldsymbol x_t; \sqrt{1-\beta_t} \boldsymbol x_{t-1}, \beta_t \boldsymbol I),
\label{eqn:ddpm_forward}
\end{equation}
where $\beta_t \in (0, 1)$ represents the noise variance at a given timestep $t \in \{1, \dots, T\}$. The distribution $q(\boldsymbol x_t | \boldsymbol x_0)$ can be derived in closed form from  \cref{eqn:ddpm_forward}. For sufficiently large $T$, the marginal $q(\boldsymbol x_T)$ approaches a standard normal distribution. A DDPM approximates the reverse Markov chain, iteratively transforming samples from a latent distribution $p(\boldsymbol x_T) = \mathcal{N}(\boldsymbol 0, \boldsymbol I)$ onto the data manifold by following
\begin{equation}
    p_\theta(\boldsymbol x_{0:T}) = p(\boldsymbol x_T)\prod_{t=1}^T p_\theta(\boldsymbol x_{t-1} | \boldsymbol x_{t}).
\label{eqn:ddpm_reverse}
\end{equation}
The reverse transition kernels are defined as 
\begin{equation}
    p_\theta(\boldsymbol x_{t-1} | \boldsymbol x_{t}) = \mathcal{N}(\boldsymbol x_{t-1}; \boldsymbol\mu^{(t)}_\theta(\boldsymbol x_t), \boldsymbol \Sigma^{(t)}_\theta (\boldsymbol x_t)),
\label{eqn:ddpm_reverse_transitions}
\end{equation}
where the distribution parameters are typically predicted by a time-dependent neural network with weights $\theta$. This network is trained to maximise a variational lower bound (VLB) on the log-likelihood $\mathbb{E}_{q(\boldsymbol x_0)} \left[ \log p_\theta(\boldsymbol x_0) \right]$. The form of the VLB loss depends on the parameterisation used to predict $\boldsymbol\mu^{(t)}_\theta(\boldsymbol x_t)$. For our purposes, we train a neural network $\hat{\boldsymbol x}^{(t)}_{0_\theta} (\boldsymbol x_t)$ to estimate the ``clean'' sample $\boldsymbol x_0$ given the noisy sample $\boldsymbol x_t$. This results in a denoising objective of the form
\begin{equation}
    \mathcal{L}_{\text{VLB}} = \mathbb{E}_{t, \boldsymbol x_0, \boldsymbol x_t | \boldsymbol x_0} \left[ \| \boldsymbol x_0 - \hat{\boldsymbol x}^{(t)}_{0_\theta} (\boldsymbol x_t)\|_2^2 \right],
\label{eqn:ddpm_loss}
\end{equation}
which corresponds to a weighted \cite{ho2020denoising} version of the VLB.

During inference, $\hat{\boldsymbol x}^{(t)}_{0_\theta} (\boldsymbol x_t)$ is used to compute  $\boldsymbol\mu^{(t)}_\theta(\boldsymbol x_t)$. Following \cite{ho2020denoising}, we set $\boldsymbol \Sigma^{(t)}_\theta (\boldsymbol x_t) = \sigma^2_t \boldsymbol I$ to time-dependent constants that depend on the hyperparameters $\beta_t$. Given $\boldsymbol\mu_\theta$ and $\boldsymbol \Sigma_\theta$, samples $\boldsymbol x_0 \sim p_\theta(\boldsymbol x_0)$ are obtained using ancestral sampling, as in \cref{eqn:ddpm_reverse}. DDPMs can be easily extended to sample from conditional distributions $p_\theta(\boldsymbol x_0 | \boldsymbol y)$, by passing a conditioning variable $\boldsymbol y$ to the denoising neural network such that ``clean'' sample estimates are given by $\hat{\boldsymbol x}^{(t)}_{0_\theta} (\boldsymbol x_t, \boldsymbol y)$.

\paragraph{Neural Implicit Surfaces.}
\label{subsec:method_background_implicit_surfaces}
A surface $\mathcal{S}$ in $\mathbb{R}^3$ can be implicitly defined as the zero-level-set or decision boundary of a function. Given an RGB image $\mathbf{I}$ of a subject, an estimate of the surface geometry of the subject may be obtained using an image-conditioned signed distance function (SDF). This can be represented by a coordinate-based neural network $f_{\Theta}$, which outputs a signed distance value $d_{\boldsymbol p}$ and unshaded albedo colour $\boldsymbol a_{\boldsymbol p}$ given a query point $\boldsymbol p \in \mathbb{R}^3$. Hereafter, we use $\Theta$ to denote the set of all learnable parameters. The neural implicit surface corresponding to $f_{\Theta}$ is defined as
\begin{equation}
    \mathcal{S}_\Theta(\mathbf{I}) 
    = \left\{ \boldsymbol p \in \mathbb{R}^3 | f_\Theta\left(\boldsymbol p; g_\Theta \left( \mathbf{I} \right) \right)
    = (0, \boldsymbol a_{\boldsymbol p}) \right\},
\label{eqn:implicit_surface}
\end{equation}
where $g_\Theta$ is a feature extractor CNN that is used to condition $f_\Theta$ on the image $\mathbf{I}$ with pixel-aligned features $g_\Theta(\mathbf{I})$, following \cite{saito2019pifu, alldieck2022phorhum}. The feature vector associated with $\boldsymbol p$, which we denote as $\boldsymbol g_{\boldsymbol p}$, is obtained by projecting $\boldsymbol p$ onto the image plane and bilinearly interpolating $g_\Theta(\mathbf{I})$ at this pixel location. The distance value $d_{\boldsymbol p}$ and albedo colour $\boldsymbol a_{\boldsymbol p}$ at $\boldsymbol p$ are concretely obtained as $ (d_{\boldsymbol p}, \boldsymbol a_{\boldsymbol p}) = f_\Theta(\boldsymbol p, \boldsymbol g_{\boldsymbol p})$.

To decouple unshaded albedo colour and illumination-dependent shading, an additional neural network $s_\Theta$ \cite{alldieck2022phorhum} may be used to estimate a shading coefficient $\boldsymbol s_{\boldsymbol p}$ at each surface point $\boldsymbol p$. This is obtained using
\begin{equation}
    \boldsymbol s_{\boldsymbol p} = s_\Theta \left( \boldsymbol n_{\boldsymbol p}, \boldsymbol l(\mathbf{I}) \right),
\label{eqn:shading}
\end{equation}
where $\boldsymbol n_{\boldsymbol p} = \nabla_{\boldsymbol p} d_{\boldsymbol p}$ is the surface normal at $\boldsymbol p$ and $\boldsymbol l(\mathbf{I})$ is a scene illumination code estimated from the input image. The latter may be computed using the bottleneck of $g_\Theta(\mathbf{I})$, as in \cite{alldieck2022phorhum}. Then, the shaded colour at a point $\boldsymbol p$ is given by $\boldsymbol c_{\boldsymbol p} = \boldsymbol a_{\boldsymbol p} \odot \boldsymbol s_{\boldsymbol p}$, where $\odot$ denotes element-wise multiplication.

Given an image-conditioned SDF $f_\Theta$, various methods exist to extract and/or render the corresponding surface $\mathcal{S}$. An explicit mesh approximation of $\mathcal{S}$ is typically generated by running Marching Cubes \cite{marching_cubes} in a densely sampled 3D bounding box \cite{saito2019pifu, saito2020pifuhd, alldieck2022phorhum}. Standard graphics pipelines can be used to render various properties of the mesh, such as surface albedo, shaded colour, surface normal or depth maps. In addition, $\mathcal{S}$ may be directly rendered using sphere tracing \cite{hart1996sphere},  without generating an explicit mesh. Sphere tracing can be formulated as a differentiable operation \cite{yariv2020multiview}, enabling the use of 2D rendering losses during training.

\textbf{%
\begin{figure*}[tp]
\vspace{-4.5mm}
    \centering
    \includegraphics[width=\linewidth,trim={0 3.2cm 0 0},clip]{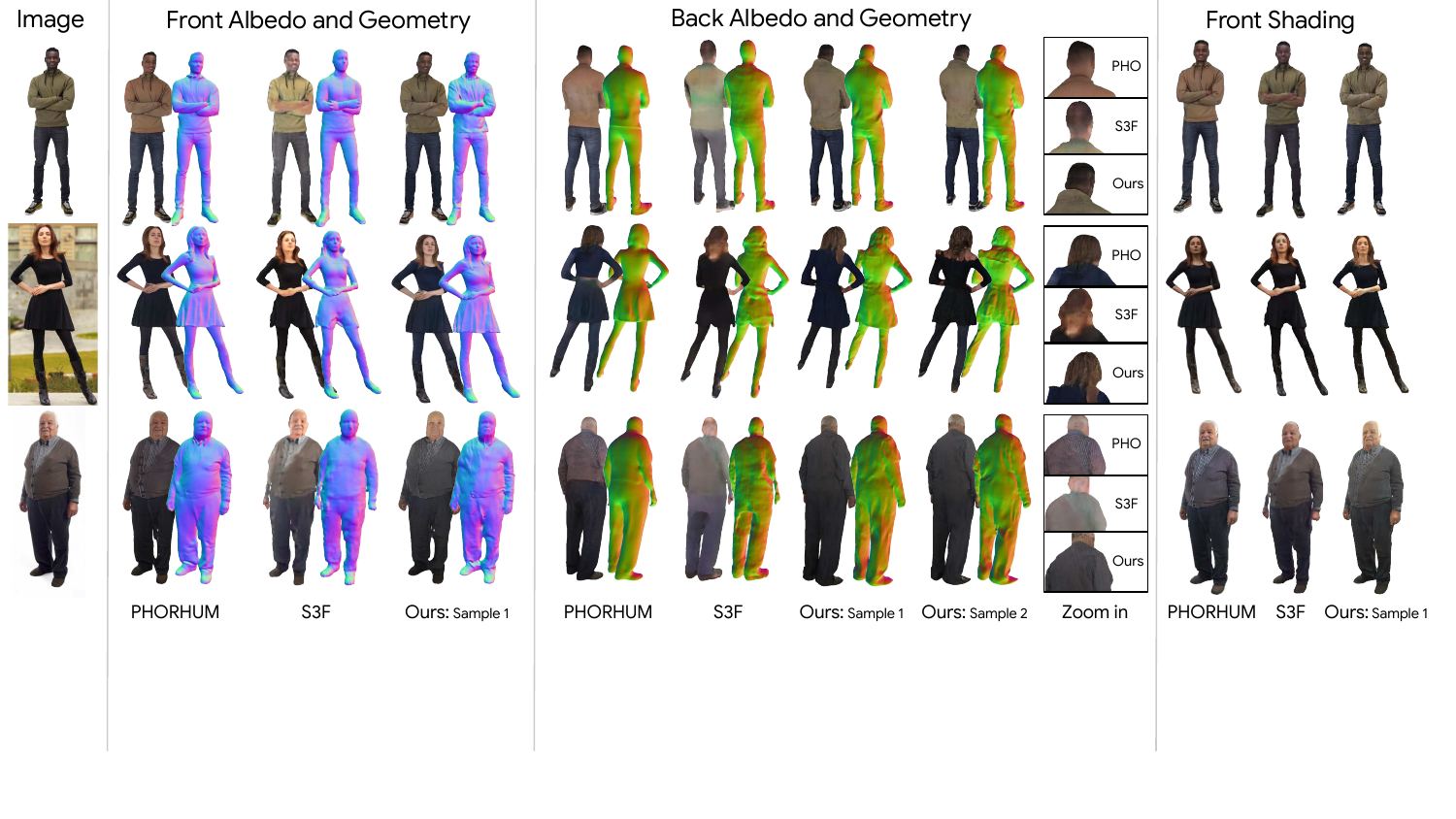}
    \caption{\textbf{Qualitative comparison against deterministic monocular 3D human reconstruction methods \cite{alldieck2022phorhum, corona2023s3f} that predict geometry, surface albedo and shaded colour.} PHORHUM \cite{alldieck2022phorhum} (retrained on our dataset) outputs good front predictions, but exhibits over-smooth, flat geometry and blurry colours on the back. S3F \cite{corona2023s3f} yields more detailed geometry, but colours are still often blurry. Moreover, both these methods occasionally paste the front colour predictions onto the back incorrectly (see row 3). Our method outputs \emph{multiple} diverse samples, with a greater level of geometric detail and colour sharpness in uncertain regions, that are consistent with the input image after shading.}
    \label{fig:compare_color}
    \vspace{-2mm}
\end{figure*}
}

\subsection{Implicit Surface Diffusion via Rendering}
\label{subsec:diffusion_via_rendering}
Our aim is to estimate the 3D surface geometry and appearance $\mathcal{S}$ of a human subject given a single RGB image $\mathbf{I}$. This is an ill-posed problem, since multiple 3D reconstructions can plausibly explain a 2D input image, \eg due to occlusion or depth ambiguity.
Thus, we seek to predict a \emph{probability distribution} over 3D geometry and appearance conditioned on the RGB image, $p_\Theta(\mathcal{S} | \mathbf{I})$. Our method, \textbf{DiffHuman}, implements $p_\Theta(\mathcal{S} | \mathbf{I})$ using the framework of DDPMs \cite{ho2020denoising}, and enables us to sample multiple plausible 3D reconstructions.

We represent $\mathcal{S}$ as an implicit surface $\mathcal{S}_\Theta(\mathbf{I})$ defined by the corresponding neural networks $f_\Theta$ and $g_\Theta$, as detailed in \cref{eqn:implicit_surface}. 
If we wish to directly apply a DDPM to learn $p_\Theta(\mathcal{S} | \mathbf{I})$, we need to define forward and reverse diffusion processes over $\mathcal{S}$. This requires a suitable representation of $\mathcal{S}$ that can be noised and denoised. In the framework of implicit surfaces, the neural network $f_\Theta$ is fixed and acts as a decoder for the pixel-aligned features $g_\Theta(\mathbf{I})$. The latter could be an adequate choice for a representation; however, they are unknown before the network is trained. Specifically, we do not have access to ground-truth pixel aligned features $g_\Theta(\mathbf{I})$ for a given $\mathbf{I}$ \textit{a priori}, and thus cannot add noise to or denoise them directly.

Instead, we model a distribution over image-based, pixel-aligned \emph{observations} of $\mathcal{S}$ that cover the true $\mathcal{S}$ well.
Specifically, we consider three types of observations of the front and back of $\mathcal{S}$ with respect to a fixed camera $\pi$: (i) unshaded albedo colour images $\mathbf{A}^F$ and $\mathbf{A}^B$, (ii) surface normal images $\mathbf{N}^F$ and $\mathbf{N}^B$, and (iii) depth maps $\mathbf{D}^F$ and $\mathbf{D}^B$. These are concatenated together to form an observation set 
\begin{equation}
    \boldsymbol x_0 = \{\mathbf{A}^F, \mathbf{A}^B, \mathbf{N}^F, \mathbf{N}^B, \mathbf{D}^F, \mathbf{D}^B \}. 
\label{eqn:x0_definition}
\end{equation}
In practice, this observation set is represented as an array $\boldsymbol x_0 \in [-1, 1]^{H \times W \times C}$. Given a surface $\mathcal{S}$, the corresponding $\boldsymbol x_0 = \mathtt{render}(\mathcal{S}, \pi)$ can be obtained via rendering. %
Since $\boldsymbol x_0$ is effectively a multichannel image, a conventional image-based DDPM may be directly applied to learn $p_\Theta(\boldsymbol x_0 | \mathbf{I})$. This would involve training a neural network $\hat{\boldsymbol x}^{(t)}_{0_\Theta} (\boldsymbol x_t, \mathbf{I})$ to estimate the ``clean'' observation set $\boldsymbol x_0$ by denoising $\boldsymbol x_t$, as detailed in \cref{subsec:method_background_ddpm}. We modify this denoising step by incorporating a neural implicit surface as an intermediate 3D reconstruction and obtain the denoised estimate via rendering of this surface. This enables us to sample from a learned distribution over 3D surfaces during inference.

Specifically, given a noisy observation set $\boldsymbol x_t$ and conditioning image $\mathbf{I}$, we first compute noise-dependent pixel-aligned features $g^{(t)}_\Theta(\boldsymbol x_t, \mathbf I)$. These are used to condition an SDF $f^{(t)}_\Theta(\boldsymbol p; g^{(t)}_\Theta(\boldsymbol x_t, \mathbf I))$, which is dependent on both $\boldsymbol x_t$ and $\mathbf{I}$. The networks $f^{(t)}_\Theta$ and $g^{(t)}_\Theta$ define a neural implicit surface $\mathcal{S}^{(t)}_\Theta(\mathbf{\boldsymbol x_t, I})$, which is given by adding noise-dependence to \cref{eqn:implicit_surface}. Then, we obtain an estimate of the denoised observation set with
\begin{equation}
    \hat{\boldsymbol x}^{(t)}_{0_\Theta} (\boldsymbol x_t, \mathbf{I}) 
    = \mathtt{render} \left( \mathcal{S}^{(t)}_\Theta(\boldsymbol x_t, \mathbf{I}), \pi \right).
\label{eqn:x0_rendering}
\end{equation}
Furthermore, we can obtain a shaded image $\mathbf{C}^{(t)}$ by applying a timestep-dependent shading network $s_\Theta^{(t)}$ to the front albedo and normal predictions that comprise $\hat{\boldsymbol x}^{(t)}_{0_\Theta}$:
\begin{equation}
\mathbf{C}^{(t)} = \mathbf{A}^F \odot s_\Theta^{(t)}(\mathbf{N}^F, \boldsymbol l(\mathbf{I})).
\label{eqn:shaded_image_ddpm}
\end{equation}
$f^{(t)}_\Theta$ and $g^{(t)}_\Theta$ are trained by minimising the following DDPM loss in each training iteration:
\begin{equation}
    \mathcal{L}^\mathtt{render}_{\text{VLB}} = \left\| \boldsymbol x_0 - \mathtt{render} \left( \mathcal{S}^{(t)}_\Theta(\boldsymbol x_t, \mathbf{I}), \pi \right) \right\|_2^2,
\label{eqn:ddpm_render_loss}
\end{equation}
which follows from \cref{eqn:ddpm_loss}. Images and corresponding clean observation sets are sampled from a target data distribution $\boldsymbol x_0, \mathbf{I} \sim q(\boldsymbol x_0, \mathbf{I})$ and a timestep is sampled from $t \sim \mathcal{U}(\{1, ..., T\})$. Noised observations are sampled from $\boldsymbol x_t \sim q(\boldsymbol x_t | \boldsymbol x_0)$.
Moreover, we can supervise on $\mathbf{C}^{(t)}$ to ensure that all 3D samples from our predicted distribution are consistent with the conditioning image after rendering and shading. Please refer to the \supp for details.

Once $f^{(t)}_\Theta$ and $g^{(t)}_\Theta$ are trained, we can ancestrally sample reverse process trajectories over observation sets $\boldsymbol x_{0:T} \sim p_\Theta(\boldsymbol x_{0:T} | \mathbf{I})$, by computing and rendering $\mathcal{S}^{(t)}_\Theta(\mathbf{\boldsymbol x_t, I})$ in each denoising step. Notably, 3D samples $\mathcal{S} \sim p_\Theta(\mathcal{S}| \mathbf{I})$ are given by the final reconstruction  $\mathcal{S} = \mathcal{S}_\Theta^{(1)}(\boldsymbol x_1, \mathbf I)$.

The above formulation of diffusion via rendering is similar to \cite{tewari2023diffusion, szymanowicz23viewset_diffusion}. These approaches implement diffusion over multiple images of an underlying 3D scene from different views, and incorporate rendering of an intermediate volumetric 3D representation \cite{nerf} into the reverse process. Our method considers various pixel-aligned observations of a 3D human from the same view and reconstructs intermediate implicit surfaces during denoising. We also incorporate probabilistic scene illumination estimation via Eqn. \ref{eqn:shaded_image_ddpm}. Nonetheless, all these approaches involve rendering of a neural 3D representation in every single denoising step, which is computationally expensive during inference. Consequently,  \cref{subsec:hybrid_diffusion} introduces a hybrid diffusion model that integrates both rendering and learned generation in the denoising process, enabling 3D sampling at considerably reduced runtime.

\subsection{Hybrid Implicit Surface Diffusion}
\label{subsec:hybrid_diffusion}

The diffusion-via-rendering formulation introduced in  \cref{subsec:diffusion_via_rendering} involves rendering an implicit surface in every denoising step. This is memory- and time-intensive, both when the surface is directly rendered using sphere tracing, and also when an explicit mesh is extracted with Marching Cubes and subsequently rasterised. Sphere tracing requires $K$ successive evaluations of $f^{(t)}_\Theta$ \emph{per pixel}, where $K$ equals the number of tracing steps until a surface is found or the ray is terminated; $K \approx 30$ in our experiments resulting in 7.9M function evaluations for an image of $512 \times 512$px. Marching Cubes, on the other hand, requires one $f^{(t)}_\Theta$ evaluation \emph{per 3D grid point}. This can be accelerated via octree sampling, but still requires $> 10^5$ function evaluations for a mesh of medium spatial resolution.
\textbf{%
\begin{figure*}[tp]
    \vspace{-6mm}
    \centering
    \includegraphics[width=\linewidth,trim={0 3.7cm 0 0},clip]{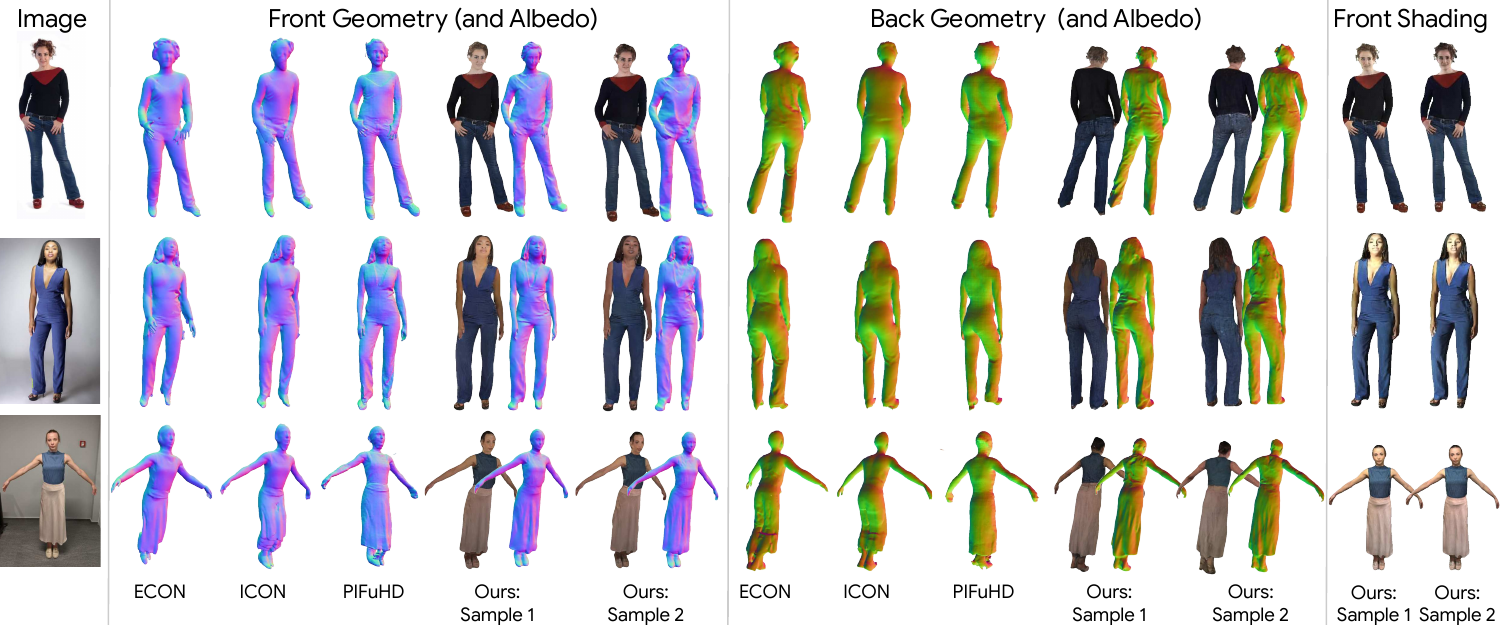}
    \caption{\textbf{Qualitative comparison against deterministic monocular 3D human reconstruction methods that predict only surface geometry: PIFuHD \cite{saito2020pifuhd}, ICON \cite{xiu2022icon} and ECON \cite{xiu2023econ}.} Samples from our method generally exhibit greater geometric detail in uncertain regions, while maintaining a high level of consistency with the input image in shaded renders. Moreover, deterministic methods often fall back towards the mean of the training data distribution when faced with ambiguous and challenging inputs \cite{bishop94mixturedensity, cui2020learning, mathieu2015deep}; \eg predicting trousers from the back instead of a long skirt in row 3. This can be mitigated by learning to predict a distribution over reconstructions instead.}
    \label{fig:compare_nocolor}
    \vspace{-3mm}
\end{figure*}
}

To mitigate this computational overhead, we note that we are ultimately only interested in 3D reconstruction samples obtained at the end of the denoising process $\mathcal{S} = \mathcal{S}_\Theta^{(1)}(\boldsymbol x_1, \mathbf I)$. Thus, explicitly computing and rendering $\mathcal{S}^{(t)}_\Theta$ in every denoising step during inference is wasteful. Instead, we introduce an additional ``generator'' neural network $h^{(t)}_\Theta$ that is trained to imitate the rendering of an implicit surface conditioned on pixel-aligned features. Concretely, $h^{(t)}_\Theta$ directly maps noise-dependent pixel-aligned features $g^{(t)}_\Theta(\boldsymbol x_t, \mathbf I)$ to an estimate of the denoised observation set 
\begin{equation}
    \bar{\boldsymbol x}^{(t)}_{0_\Theta} (\boldsymbol x_t, \mathbf{I}) 
    = h_\Theta^{(t)} \left( g^{(t)}_\Theta(\boldsymbol x_t, \mathbf I)\right),
\label{eqn:x0_decoding}
\end{equation}
where $\bar{\boldsymbol x}^{(t)}_{0_\Theta}$ should approximate $\hat{\boldsymbol x}^{(t)}_{0_\Theta}$ obtained via rendering (\cref{eqn:x0_rendering}). $h^{(t)}_\Theta$ is trained with the following objective: 
\begin{equation}
    \mathcal{L}^\mathtt{generate}_{\text{VLB}} = \left\| \boldsymbol x_0 - h_\Theta^{(t)} \left( g^{(t)}_\Theta(\boldsymbol x_t, \mathbf I)\right) \right\|_2^2.
\label{eqn:ddpm_decode_loss}
\end{equation}

During inference, we sample reverse process trajectories $\boldsymbol x_{1:T} \sim p_\Theta(\boldsymbol x_{1:T} | \mathbf{I})$ using generated denoised estimates $\bar{\boldsymbol x}^{(t)}_{0_\Theta}$ instead of rendering. We only explicitly compute 3D reconstruction samples $\mathcal{S} = \mathcal{S}_\Theta^{(1)}(\boldsymbol x_1, \mathbf I)$ at the end of the reverse process, using the final noisy observations $\boldsymbol x_1$. Marching Cubes is simply applied once to extract the final mesh. In general, a forward pass through the neural network $h^{(t)}_\Theta$ is computationally cheaper than explicit rendering via sphere tracing or Marching Cubes and rasterisation. The computational cost savings sum together over the reverse process, which may involve hundreds of denoising steps.

Generating denoised estimates with the composition of neural networks $h^{(t)}_\Theta \circ g^{(t)}_\Theta$ is reminiscent of the standard denoising network architecture used in conventional DDPMs. However, we simultaneously apply both $\mathcal{L}^\mathtt{render}_{\text{VLB}}$ and $\mathcal{L}^\mathtt{generate}_{\text{VLB}}$ during training, resulting in a hybrid diffusion framework that combines rendering and generation. This ensures that $g^{(t)}_\Theta(\boldsymbol x_t, \mathbf I)$ continue to be features that validly condition an SDF and $h^{(t)}_\Theta$ learns to decode these features into the observation sets corresponding to the SDF.

\subsection{Implementation Details}
\label{subsec:method_impl_details}
Our networks are trained with a synthetic training dataset, consisting of HDRI-based illuminated renders of real body scans from \cite{renderpeople} and our own captured data.
We use $\sim5.9$K scans of $\sim1.1$K identities to render $\sim450$K training examples, each consisting of a $512 \times 512$px image (with masked-out background) and an observation set $\boldsymbol x_0$.
The back-views in $\boldsymbol x_0$ are created by inverting the z-buffer, and thus rendering the scans back-to-front. \cite{renderpeople} provides ground-truth albedo textures, which we use to generate $\mathbf{A}^F$ and $\mathbf{A}^B$ in $\boldsymbol x_0$. Our scans approximately capture albedo using even ambient lighting. However, this is not perfect and causes our model to sometimes yield shading artefacts in albedo predictions.

In addition to $\mathcal{L}_{\text{VLB}}^*$, we use 3D losses on $d_{\boldsymbol p}$, $\boldsymbol a_{\boldsymbol p}$ and $\boldsymbol n_{\boldsymbol p}$ to improve training stability (see \supp for details). 
During training, we render $32 \times 32$px patches using differentiable ray tracing \cite{yariv2020multiview} to form $\hat{\boldsymbol x}^{(t)}_{0_\Theta}$ for  $\mathcal{L}_{\text{VLB}}^\mathtt{render}$. We apply $\mathcal{L}_{\text{VLB}}^\mathtt{generate}$ on the full resolution generation $\bar{\boldsymbol x}^{(t)}_{0_\Theta}$.
$g_\Theta$ and $h_\Theta$ are U-Nets \cite{unet} with 13 encoder-decoder layers each and skip connections. Both networks double the filter size in each encoder layer, starting from $64$ up to $512$ for $g_\Theta$ and up to $128$ for $h_\Theta$. $g_\Theta$ outputs a pixel-aligned feature map in $\mathbb{R}^{512 \times 512 \times 256}$.
$f_\Theta$ and $s_\Theta$ are MLPs, following \cite{alldieck2022phorhum}.

At test time, we perform 100 DDIM \cite{song2020denoising} denoising steps. At each step, we can choose to denoise via $\mathtt{render}$ or $\mathtt{generate}$ and we ablate different strategies in \cref{subsec:ablations}. For faster inference, we use Marching Cubes and rasterisation, instead of sphere tracing, in $\mathtt{render}$. If we $\mathtt{render}$ at higher noise (large $t$), we run Marching Cubes on a $256^3$ grid. For small $t$, we use $512^3$.
The final denoising step always has to be a $\mathtt{render}$ step, since $\mathtt{generate}$ does not produce 3D geometry.
However, we do not perform a full step of $\mathtt{render}$ at $t=1$, but only reconstruct $\mathcal{S}$ and omit rasterisation of $\boldsymbol x_0$.

\begin{figure*}[tp]
   \vspace{-5mm}
    \centering
    \includegraphics[width=0.98\linewidth,trim={0 8.9cm 0 0},clip]{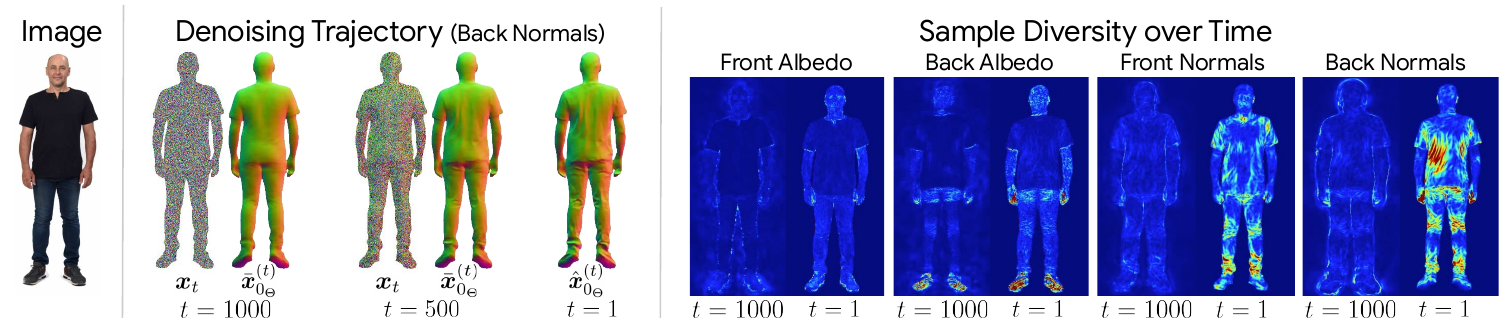}
    \caption{
    \textbf{Visualisation of the reverse process.} The denoising trajectory shows noisy samples $\boldsymbol x_t$ and generated clean predictions $\bar{\boldsymbol x}^{(t)}_{0_\Theta}$ at each timestep. Clean predictions are initially very simple, akin to many deterministic approaches, and become detailed over time. The heatmaps show sample diversity, computed as the per-pixel variance of the observations in $\bar{\boldsymbol x}^{(t)}_{0_\Theta}$ over 10 samples. Diversity is low at the start of the denoising process ($t = 1000$), but increases gradually as the samples diverge. Back diversity is, intuitively, greater than the front.
    }
\vspace{-0.3cm}
\label{fig:diversity}
\end{figure*}

\section{Experiments}
\label{sec:implementation_details}
This section quantitatively compares \method with the state-of-the-art photorealistic human reconstruction methods, and visually demonstrates the quality of 3D samples conditioned on internet images.
Furthermore, we experimentally ablate a number of crucial design choices.
Please see the \supp for additional results and experiments.

\paragraph{Test dataset and metrics.}
We use the test set of \cite{alldieck2022phorhum} for numerical evaluation, and report both 3D metrics and image-based (pixel-aligned) metrics.

The 3D metrics consist of bi-directional Chamfer distance $\times 10^{-3}$ (Ch.~$\downarrow$), Normal Consistency (NC $\uparrow$), and Volumetric Intersection over Union (IoU $\uparrow$). 
Iterative Closest Points is used to first align 3D predictions with the ground-truth. 

3D metrics are sensitive to the assumed camera model.
In contrast, image-based metrics partially ignore errors due to an incorrect camera, instead focusing on surface structure. This is better correlated with perceived quality.
Image-based metrics are computed by rendering 3D predictions  with each model's assumed camera, and comparing the resulting images against ground-truth 2D renders.
Specifically, we report Structural Similarity Index (SSIM $\uparrow$), Learned Perceptual Image Patch Similarity (LPIPS $\downarrow$) \cite{zhang2018perceptual}, and Peak Signal-to-Noise Ratio (PSNR $\uparrow$) for albedo and shaded colour renders.
For normal renders, we report the angular error in degrees (Ang. $\downarrow$) and LPIPS.
SSIM and LPIPS evaluate structure rather than pixel-to-pixel errors.
The latter can be misleadingly low for over-smooth reconstructions; SSIM and LPIPS, in our experience, better capture the perceived quality.

For our method, we report metrics using the best-of-$N \in \{1, 5, 10\}$ reconstructions, following \cite{biggs20203d, kolotouros2021probabilistic, sengupta2023humaniflow}.
Specifically, we obtain $N$ different 3D samples for each test image, and aggregate metrics using the numerical best reconstruction. 
The \supp additionally reports mean metrics, and discusses the use of best-of-$N$ vs.\ mean metrics for this task.

\paragraph{Baselines.}
We compare with a large number of recent approaches to monocular 3D human reconstruction.
Only PHORHUM \cite{alldieck2022phorhum}, S3F \cite{corona2023s3f}, and PIFu \cite{saito2019pifu} reconstruct surface color, but the latter does not decompose albedo and shading.
ARCH \cite{huang2020arch} and ARCH++ \cite{he2021arch++} do not reconstruct true surface details but use normal mapping to enhance the visual fidelity of results.
For fairness, we evaluate the estimated normals instead of true surface normals for these methods.
We also compare against a version of PHORHUM retrained with our larger dataset, resulting in a strong baseline method.

\begin{table}[t]
\centering
\resizebox{\columnwidth}{!}{%
\renewcommand{\tabcolsep}{3.5pt}

\begin{tabular}{l | c | c c c c}

\multirow{2}{0.2\columnwidth}{Render frequency} & Runtime & 3D & Alb.\ F / B & Nor.\ F / B & Sha.\ F\\
& s / sample & Ch.\ $\downarrow$ & PSNR $\uparrow$ & Ang.\ $\downarrow$ & PSNR $\uparrow$ \\
\hline
\hline
Per step & 496 & \cbest{0.99} & 22.92 / 20.95 & 21.59 / 22.88 & 26.57 \\
Per 10 steps & 86 & \csecond{1.12} & 23.24 / 21.07 & \cbest{19.05 / 22.46} & \csecond{27.08}\\
Per 25 steps & 34 & 1.16 & \cbest{23.26 / 21.06} & \csecond{19.11 / 22.52} & \cbest{27.09}\\
Final step & 9 & 1.16 & \csecond{23.26 / 21.05} & 19.12 / 22.55 & \cbest{27.09} \\
\end{tabular}
}
\caption{
\textbf{Ablation of hybrid implicit surface diffusion.} $N=5$ samples are obtained using 100 DDIM \cite{song2020denoising} steps. We periodically $\mathtt{render}$ every 1, 10 or 25 steps, or only in the final step. All other denoising steps use $\mathtt{generate}$. While per-step $\mathtt{render}$ performs best on 3D metrics, predominantly using $\mathtt{generate}$ results in better colour. The \best{best} and \second{second best} results are marked.
}
\vspace{-0.05in}
\label{tab:ablation_gen_vs_rend}
\end{table}

\subsection{Ablation Studies}
\label{subsec:ablations}

Since we retrained PHORHUM \cite{alldieck2022phorhum} on our larger synthetic dataset, we consider it as an ablation of our main design choice: probabilistically modelling the reconstruction process using a diffusion model that predicts distributions over 3D human reconstructions.
Even though the retrained PHORHUM model turned out to be a very strong baseline, \method is able to produce reconstructions with higher visual fidelity and better numerical performance, especially for unseen regions.
Better performance for unseen regions can be explained by our probabilistic approach being less prone to averaging effects caused by the inherent aleatoric \cite{kendall2017uncertainties} uncertainty in an ill-posed problem.

Additionally, we ablate our hybrid diffusion strategy: the generator network $h_\Theta^{(t)}$ and denoising via $\mathtt{generate}$ instead of $\mathtt{render}$.
In \cref{tab:ablation_gen_vs_rend} we compare the use of $\mathtt{render}$ in every denoising step, every 10, every 25, and only at the final step (to extract the final mesh with Marching Cubes). We use $\mathtt{generate}$ for all other steps. The performances of all variants are comparable, suggesting that $h^{(t)}_\Theta$ has learned to imitate $\mathtt{render}$ sufficiently well.
Nonetheless, Ch.\ is slightly better when using $\mathtt{render}$ in every step, while colour metrics are improved with lower $\mathtt{render}$ frequency.
This is expected: $\mathtt{render}$ actually reconstructs 3D geometry whereas $\mathtt{generate}$ only synthesises observations. 
Errors in this approximate synthesis operation may accumulate over the course of the denoising process.
On the other hand, $\mathtt{generate}$ may also ``fix'' inconsistent colour extracted from the signed-distance and colour field.
Crucially, the use of $\mathtt{generate}$ results in an up to $55\times$ speed up at comparable quality.
We use the ``final step'' strategy for all remaining experiments.
The \supp shows that 3D samples obtained via per-step and final step rendering are visually similar.

\begin{table*}[t]
   \vspace{-4mm}
\centering
\resizebox{\linewidth}{!}{

\begin{tabular}{l | c c c | c c c | c c | c c | c c c }
\multirow{2}{*}{\textbf{Method}} & \multicolumn{3}{c|}{\textbf{Albedo Front}} & \multicolumn{3}{c|}{\textbf{Albedo Back}} & \multicolumn{2}{c|}{\textbf{Normals Front}} & \multicolumn{2}{c|}{\textbf{Normals Back}} & \multicolumn{3}{c}{\textbf{Shaded Front}}\\
& SSIM $\uparrow$ & LPIPS $\downarrow$ & PSNR $\uparrow$ & SSIM $\uparrow$ & LPIPS $\downarrow$ & PSNR $\uparrow$ & Ang.\ $\downarrow$ & LPIPS $\downarrow$ & Ang.\ $\downarrow$ & LPIPS $\downarrow$ & SSIM $\uparrow$ & LPIPS $\downarrow$ & PSNR $\uparrow$ \\
\hline
\hline
PIFu \cite{saito2019pifu} & -- & -- & -- & -- & -- & -- & 26.69 & 0.17 & 28.49 & 0.26 & 0.83 & 0.16 & 24.57 \\
PIFuHD \cite{saito2020pifuhd} & -- & -- & -- & -- & -- & -- & 23.04 & 0.10 & 26.33 & 0.22 & -- & -- & -- \\
Geo-PiFU \cite{he2020geo} & -- & -- & -- & -- & -- & -- & 30.17 & 0.19 & 31.93 & 0.26 & -- & -- & -- \\
ARCH \cite{huang2020arch} & -- & -- & -- & -- & -- & -- & 32.20 & 0.20 & 33.96 & 0.27 & 0.72 & 0.23 & 19.28 \\
ARCH++ \cite{he2021arch++} & -- & -- & -- & -- & -- & -- & 27.20 & 0.17 & 30.62 & 0.24 & 0.83 & 0.17 & 22.69 \\
PHORHUM \cite{alldieck2022phorhum} & \csecond{0.85} & \csecond{0.12} & 22.23  & \cbest{0.76} & \csecond{0.22} & 20.19 & 20.53 & \cbest{0.11} & 23.55 & 0.20 & 0.85 & 0.13 & 24.01 \\
PaMIR \cite{zheng2021pamir} & -- & -- & -- & -- & -- & -- & 22.88 & 0.14 & 27.26 & 0.23 & -- & -- & --  \\
ICON \cite{xiu2022icon} & -- & -- & -- & -- & -- & -- & 23.57 & 0.14 & 26.98 & 0.23 & -- & -- & -- \\
ECON \cite{xiu2023econ} & -- & -- & -- & -- & -- & -- & 22.27 & 0.15 & 26.98 & 0.23 & -- & -- & -- \\
D-IF \cite{yang2023d} & -- & -- & -- & -- & -- & -- & 24.52 & 0.15 & 27.84 & 0.23 & -- & -- & --  \\
S3F \cite{corona2023s3f} & 0.60 & 0.36 & 15.33 & 0.63 & 0.39 & 15.99 & 23.76 & 0.25 & 23.72 & 0.27 & 0.61 & 0.33 & 17.38 \\
PHORHUM (retr.)  & \csecond{0.85} & \cbest{0.11} & 22.57 & 0.73 & \cbest{0.21} & 19.74 & \cbest{18.41} & \csecond{0.12} & 22.82 & 0.19 & 0.86 & \csecond{0.10} & 25.23 \\
\hline
DiffHuman: $N = 1$ & 0.84 & \csecond{0.12} & 22.44 & 0.71 & 0.23 & 19.77 & 19.70 & 0.14 & 24.34 & 0.18 & 0.89 & \csecond{0.10} & 26.81 \\
DiffHuman: $N = 5$ & \cbest{0.86} & \cbest{0.11} & \csecond{23.26} & 0.74 & \csecond{0.22} & \csecond{21.05} & 19.12 & 0.13 & \csecond{22.55} & \csecond{0.16} & \cbest{0.90} & \csecond{0.10} & \csecond{27.09} \\
DiffHuman: $N = 10$ & \cbest{0.86} & \cbest{0.11} & \cbest{23.47} & \csecond{0.75} & \cbest{0.21} & \cbest{21.24} & \csecond{18.91} & 0.13 & \cbest{22.34} & \cbest{0.15} & \cbest{0.90} & \cbest{0.09} & \cbest{27.15} \\
\end{tabular}
}
\caption{
\textbf{Quantitative comparison against other monocular 3D human reconstruction methods in terms of pixel-aligned metrics.} Since \method predicts a distribution over 3D reconstructions, we report metrics using the best of $N=1, 5$ and $10$ samples drawn for each test image. We $\mathtt{render}$ only in the final denoising step, and use $\mathtt{generate}$ otherwise. The \best{best} and \second{second best} results are marked.
}
\vspace{-0.05in}
\label{tab:render_metrics_vs_sota}
\end{table*}

\begin{table}[t]
\centering
\footnotesize

\begin{tabular}{l | c c c}
\textbf{Method} & Ch.\ $\downarrow$ & IoU $\uparrow$ & NC $\uparrow$\\
\hline
PIFu \cite{saito2019pifu} & 3.21 & 0.61 & 0.77 \\
PIFuHD \cite{saito2020pifuhd} & 4.54 & 0.62 & 0.78 \\
Geo-PIFu \cite{he2020geo} & 4.98 & 0.54 & 0.72 \\
ARCH  \cite{huang2020arch} $\dagger$ & 3.58 & 0.57 & 0.75 \\
ARCH++  \cite{he2021arch++} $\dagger$ & 3.48 & 0.59 & 0.77 \\
PaMIR  \cite{zheng2021pamir} $\dagger$ & 2.88 & 0.61 & 0.77 \\
PHORHUM \cite{alldieck2022phorhum} & 1.29 & \cbest{0.73} & 0.85 \\
ICON  \cite{xiu2022icon} $\dagger$ & 2.44 & 0.62 & 0.78 \\
ECON\cite{xiu2023econ} $\dagger$  & 3.48 & 0.61 & 0.76 \\
D-IF \cite{yang2023d} $\dagger$  & 2.97 & 0.58 & 0.78 \\
S3F \cite{corona2023s3f} $\dagger$  & 2.35 & 0.63 & 0.80 \\
PHORHUM  (retrained)  & \csecond{1.10} & \cbest{0.73} & \cbest{0.87} \\
\hline

DiffHuman: $N = 1$ & 1.98 & 0.69 & 0.83 \\
DiffHuman: $N = 5$ & 1.16 & \csecond{0.72} & \csecond{0.86} \\
DiffHuman: $N = 10$ & \cbest{1.09} & \cbest{0.73} & \csecond{0.86} \\
\end{tabular}
\caption{
\textbf{Quantitative comparison against other monocular 3D human reconstruction methods in terms of 3D metrics.} Since \method predicts a distribution over 3D reconstructions, we report metrics using the best of $N=1, 5$ and $10$ samples drawn for each test image. We $\mathtt{render}$ only in the final denoising step, and use $\mathtt{generate}$ otherwise. The \best{best} and \second{second best} results are marked. Methods marked with $\dagger$ use a parametric body model.
}
\vspace{-0.1in}
\label{tab:3d_metrics_vs_sota}
\end{table}

\subsection{Reconstruction Accuracy}
\cref{tab:render_metrics_vs_sota} and \cref{tab:3d_metrics_vs_sota} compare recent methods in terms of image-based metrics and 3D metrics respectively.
\method yields improved performance for image-based back metrics, especially with growing number of samples $N$, while front metrics are stable for all $N$.
This is because the back is unobserved -- more samples means a higher chance of finding the correct reconstruction -- while the front is visible and thus variation is lower.
We do not split observed and unobserved parts for 3D metrics, but also find a global improvement with higher $N$.
Our retrained PHORHUM also performs well, and often slightly better than $N=1$.
This is again expected, as \method is conducting a much harder task: while PHORHUM outputs a single solution, \method models the distribution over possible 3D reconstructions.
Furthermore, the 3D ground-truth is \emph{but one} plausible reconstruction in a monocular setting.
Our method is able to yield other 3D solutions that are input-consistent, but differ from the ground-truth resulting in worse metrics.
Nevertheless, \method is competitive even for $N=1$, and produces qualitatively better reconstructions, as discussed below.

\subsection{Qualitative Results and Diversity}
We show the qualitative performance of \method in \cref{fig:compare_color,fig:compare_nocolor} side-by-side with state-of-the-art approaches.
All competing methods only return one solution, while \method allows us to sample multiple diverse results. 
We show two reconstructions per image.
Consistent with the numerical results in \cref{tab:3d_metrics_vs_sota}, \method can shine the most when reconstructing unobserved back-sides.
Despite performing well numerically, our retrained PHORHUM baseline does not produce good reconstructions of uncertain regions, with blurry colours and a lack of geometric detail.
Methods that explicitly estimate a back normal map \cite{xiu2023econ, xiu2022icon, saito2020pifuhd} produce over-smooth back reconstructions.
In contrast, samples from \method exhibit fine wrinkles and details both for observed and unobserved regions, as shown by rows 1 and 2 in \cref{fig:compare_nocolor}.
PHORHUM and S3F \cite{corona2023s3f} tend to simply clone front colours to the person's back-side -- a reasonable approach for some but not all garments. \Eg in the 3rd row of \cref{fig:compare_color}, the subject's shirt is cloned onto the jacket in the back.
In contrast, \method reliably colours unobserved regions without such artefacts.
Moreover, we can obtain diverse reconstructions, shown by the different dresses and hairstyles in \cref{fig:teaser} and row 2 of \cref{fig:compare_color}.
We visualise diversity over the denoising process in \cref{fig:diversity}.
The \supp contains additional qualitative results, as well as unconditional and edge-conditioned samples from \method. 
These exhibit greater diversity than image-conditioned samples, as RGB images are strong conditioning signals.

\section{Conclusion}
\label{sec:conclusion}
We presented \method, a probabilistic method for photorealistic 3D human reconstruction from a single RGB image.
We build on top of recent advances in diffusion-based generative modelling and propose a novel pipeline for fast sampling of 3D human shapes.
Our model is numerically competitive with the state-of-the-art, while improving the visual fidelity and the level of detail of unseen surfaces.
Furthermore, we can sample multiple input-consistent but diverse 3D human reconstructions.
Our novel hybrid implicit surface diffusion speeds up 3D sampling at test time compared with diffusion-via-rendering \cite{tewari2023diffusion, szymanowicz23viewset_diffusion}, giving a general framework for computationally cheaper diffusion over implicit 3D representations.
A limitation of our method is that it currently requires examples with known 3D geometry for training, which constrains the amount of data that can be used.
In future work, we plan to overcome this by leveraging data with partial 2D and 2.5D supervision.

\ifarxiv
\clearpage
\appendix
\section*{\Large Supplementary Material}
\appendix

This supplementary material provides additional implementation details, experiments and qualitative results supporting the main manuscript. In particular, Section \ref{sec:supmat_imp_det} gives further details on the losses used to train our DiffHuman model. Section \ref{sec:supmat_ablation} provides ablation studies investigating: (i) different types of observation sets, (ii) classifier-free diffusion guidance, and (iii) diffusion-via-rendering vs.\ hybrid diffusion. Moreover, we report mean and standard deviation metrics, to complement the ``best of $N$'' metrics reported in the main paper. Finally, Section \ref{sec:supmat_quali} provides further qualitative comparisons with competing approaches, as well as some examples of unconditional 3D reconstruction samples.

\section{Implementation Details}
\label{sec:supmat_imp_det}

Our implicit surface diffusion model is trained using 2 different denoising objectives, given by Eqns.\ \ref{eqn:ddpm_render_loss} and \ref{eqn:ddpm_decode_loss} in the main manuscript. In addition, we employ a probabilistic shaded rendering loss to ensure that 3D reconstruction samples are consistent with the 2D conditioning image, as well as several 3D losses on the intermediate implicit surface to stabilise training. These are detailed below. 

\nbf{Shaded rendering loss} We want 3D reconstruction samples, represented by implicit surfaces $\mathcal{S}^{(t)}_\Theta$, to be consistent with the conditioning image $\mathbf{I}$ at every timestep $t$ in the reverse diffusion process. This is achieved by enforcing shaded front renders of $\mathcal{S}^{(t)}_\Theta$ to match $\mathbf{I}$. Shaded renders can be obtained using the albedo and surface normal images that comprise the observation sets we use for diffusion-via-rendering and hybrid diffusion. 

Recall that ground-truth observation sets consist of $\boldsymbol x_0 = \{ \mathbf{A}^F, \mathbf{A}^B, \mathbf{N}^F, \mathbf{N}^B, \mathbf{D}^F, \mathbf{D}^B \}$. The reverse process generates samples by repeatedly estimating denoised observations given noisy observations $\boldsymbol x_t$. A denoised estimate obtained using $\mathtt{render}$ is denoted as $\hat{\boldsymbol x}^{(t)}_{0_\Theta}$, while $\bar{\boldsymbol x}^{(t)}_{0_\Theta}$ represents an estimate given by $\mathtt{generate}$ (see Eqns.\ \ref{eqn:x0_rendering} and \ref{eqn:x0_decoding} in the main paper). Front albedo $\mathbf{A}^F$ and front normals $\mathbf{N}^F$ may be used in conjunction with the shading neural network $s_\Theta^{(t)}$ to obtain shaded front images $\mathbf{C}^{(t)}$ at each timestep $t$. We compute separate $\mathbf{C}^{(t)}_\mathtt{render}$ and $\mathbf{C}^{(t)}_\mathtt{generate}$ using the elements of $\hat{\boldsymbol x}^{(t)}_{0_\Theta}$ and $\bar{\boldsymbol x}^{(t)}_{0_\Theta}$, respectively. During training, we apply $L_2$ losses between shaded renders and the condition $\mathbf{I}$, given by
\begin{align}
    \mathcal{L}_\text{shaded}^\mathtt{render} & = \| \mathbf{C}^{(t)}_\mathtt{render} - \mathbf{I} \|^2_2\\
    \mathcal{L}_\text{shaded}^\mathtt{generate} & = \| \mathbf{C}^{(t)}_\mathtt{generate} - \mathbf{I} \|^2_2.
\end{align}
Note that shaded render losses $\mathcal{L}_\text{shaded}$ have a similar form to the the denoising diffusion objectives $\mathcal{L}_\text{VLB}$. However, $\mathcal{L}_\text{shaded}$ enforces consistency between an estimated observation set and the \emph{conditioning image} $\mathbf{I}$, while $\mathcal{L}_\text{VLB}$ is applied between the estimated and ground-truth observation sets.

\begin{table}[t]
\centering
\renewcommand{\arraystretch}{1.2}
\resizebox{\columnwidth}{!}{%

\begin{tabular}{l | c | c | c}
\textbf{Loss} & \textbf{Symbol} & \textbf{Type} & \textbf{Weight} \\
\hline
\hline
Denoising-via-rendering & $\mathcal{L}_\text{VLB}^\mathtt{render}$ & Probabilistic & 1.0 \\
Denoising-via-generation & $\mathcal{L}_\text{VLB}^\mathtt{generate}$ & Probabilistic & 1.0 \\
Shaded rendering & $\mathcal{L}_\text{shaded}^\mathtt{render}$ & Probabilistic & 1.0 \\
Shaded generation & $\mathcal{L}_\text{shaded}^\mathtt{generate}$ & Probabilistic & 1.0 \\
\hline
On-surface SDF on $d_{\boldsymbol p}$ & - & Deterministic & 1.0 \\
On-surface albedo on $\boldsymbol a_{\boldsymbol p}$ & - & Deterministic & 0.2 \\
On-surface normals on $\boldsymbol n_{\boldsymbol p}$ & - & Deterministic & 0.2 \\
Near-surface In/Out on $d_{\boldsymbol p}$ & - & Deterministic & 0.2 \\
Near-surface albedo on $a_{\boldsymbol p}$ & - & Deterministic & 0.2 \\
\hline
Eikonal & - & Regulariser & 0.05 \\

\end{tabular}

}
\caption{
Summary of the probabilistic, deterministic and regularisation losses used to train our model. Loss weights are provided.
}
\label{tab:supmat_losses}
\end{table}

\nbf{Deterministic 3D losses}
$\mathcal{L}_\text{VLB}$ and $\mathcal{L}_\text{shaded}$ are probabilistic losses applied within the diffusion framework. In addition, we employ several deterministic 3D losses on surface geometry and albedo, following PIFu \cite{saito2019pifu} and PHORHUM \cite{alldieck2022phorhum}, which improves training stability in our experience. 

Specifically, we supervise SDF values $d_{\boldsymbol p}$, albedo field values $\boldsymbol a_{\boldsymbol p}$ and per-point normals $\boldsymbol n_{\boldsymbol p}$ at 3D points $\boldsymbol p$ sampled from the ground-truth 3D human surface. $d_{\boldsymbol p}$ is enforced to be $0$ for these on-surface points. Additionally, we supervise the sign of samples taken around the surface using an inside-outside classification loss implemented using binary cross-entropy. This is applied to the SDF values $d_{\boldsymbol p}$ for near-surface points. Moreover, following \cite{saito2019pifu}, the albedo colour field $\boldsymbol a_{\boldsymbol p}$ is also supervised for near-surface points. The ground-truth near-surface albedo at $\boldsymbol p$ is approximated using the albedo of the nearest neighbour on the ground-truth surface. Finally, we use an Eikonal geometric regulariser \cite{eikonal_loss} to enforce SDF predictions to have unit norm gradients everywhere.

All losses are summarised in \cref{tab:supmat_losses} in this supplementary material, which also provides associated loss weight hyperparameters. Note that deterministic losses are generally weighted lower than probabilistic losses, to encourage sample diversity. Future work may investigate the feasibility of removing deterministic losses altogether. 

\begin{table*}[t]
\centering
\resizebox{\linewidth}{!}{
\renewcommand{\tabcolsep}{4pt}
\begin{tabular}{ l | l | c | c c | c c | c c | c | c | c}
&
\multirow{2}{0.1\columnwidth}{\textbf{Render Freq.}} &
\textbf{Runtime} &
\multicolumn{2}{c|}{\textbf{3D}} & 
\multicolumn{2}{c|}{\textbf{Albedo Front}} & 
\multicolumn{2}{c|}{\textbf{Albedo Back}} &
\textbf{Normals Front} &
\textbf{Normals Back} &
\textbf{Shaded Front}\\
&
&
s / sample &
CD $\downarrow$  & NC $\uparrow$ &
LPIPS $\downarrow$ & PSNR $\uparrow$ &
LPIPS $\downarrow$ & PSNR $\uparrow$ & 
Ang.\ $\downarrow$ & 
Ang.\ $\downarrow$ & 
PSNR $\uparrow$ \\
\hline
\hline
\multirow{4}{0.15\columnwidth}{Best of $N = 5$} & 
Per step & 
496 & \cbest{0.99} & 0.85 & 0.13 & 22.92 & 0.25 & 20.95 & 21.59 & 22.88  & 26.57 \\
& Per 10 & 
86 & \csecond{1.12} & \cbest{0.87} &  0.12 & 23.24 & 0.24 & \cbest{21.07} & \cbest{19.05} & \cbest{22.46}  & \csecond{27.08} \\
& Per 25 & 
\csecond{34} & 1.16 & \csecond{0.86} & \csecond{0.11} & \cbest{23.26} & \csecond{0.23} & \csecond{21.06} & \csecond{19.11} & \csecond{22.52}  & 27.09 \\
& Final & 
\cbest{9} & 1.16 & 0.86 & \cbest{0.11} & \csecond{23.26} & \cbest{0.22} & 21.05 & 19.12 & 22.55 & \cbest{27.09}\\
\hline
\multirow{4}{0.15\columnwidth}{Mean $\pm$ Std. $N = 5$} & 
Per step & 
496 & \cbest{1.03} $\pm$ 0.8 & 0.83 $\pm$ 0.04 & 0.14 $\pm$ 0.03 & 22.34 $\pm$ 2.39 & 0.27 $\pm$ 0.07 & 20.34 $\pm$ 3.35 & 22.47 $\pm$ 3.13  & 23.42 $\pm$ 5.49  & 26.18 $\pm$ 1.90\\
& Per 10 & 
86 & \csecond{1.33} $\pm$ 0.9 & \cbest{0.85} $\pm$ 0.05 & 0.13 $\pm$ 0.04 & 22.63 $\pm$ 2.40 & 0.24 $\pm$ 0.08 & \cbest{20.46} $\pm$ 3.36 & \cbest{20.14} $\pm$ 3.20  & \cbest{23.15} $\pm$ 5.64  & 26.82 $\pm$ 1.91 \\
& Per 25 & 
\csecond{34} & 1.37 $\pm$ 0.9 & \csecond{0.84} $\pm$ 0.05 & \csecond{0.13} $\pm$ 0.04 & \csecond{22.64} $\pm$ 2.43 & \csecond{0.24} $\pm$ 0.08 & 20.45 $\pm$ 3.38 & \csecond{20.20} $\pm$ 3.16  & 23.31 $\pm$ 5.56 & \csecond{26.83} $\pm$ 1.90 \\
& Final & 
\cbest{9} & 1.38 $\pm$ 0.9 & 0.84 $\pm$ 0.05 & \cbest{0.12} $\pm$ 0.04 & \cbest{22.65} $\pm$ 2.43 & \cbest{0.23} $\pm$ 0.08 & \csecond{20.46} $\pm$ 3.37 & 20.24 $\pm$ 3.14  & \csecond{23.20} $\pm$ 5.54 & \cbest{26.84} $\pm$ 1.90

\end{tabular}
}
\caption{
Ablation of hybrid implicit surface diffusion. $N=5$ samples are obtained using 100 DDIM \cite{song2020denoising} steps. We ablate periodic $\mathtt{render}$ every 1, 10 and 25 steps, as well as only in the final step. The latter only involves running Marching Cubes \cite{marching_cubes} for mesh extraction in the final step. All other denoising steps use $\mathtt{generate}$. While per step $\mathtt{render}$ performs best on 3D metrics, predominantly using $\mathtt{generate}$ results in better perceptual quality in rendered metrics. The \best{best} and \second{second best} results are marked. We report both best-of-$N$ and mean ($\pm$ std.) metrics for completeness.
}

\label{tab:supmat_gen_vs_rend}
\end{table*}

\begin{table*}[t]
\centering
\resizebox{\linewidth}{!}{
\renewcommand{\tabcolsep}{4pt}
\begin{tabular}{ l | l l | c c | c c | c c | c | c | c}
&
\multicolumn{2}{l|}{\textbf{Observations in $\boldsymbol x_0$}} &
\multicolumn{2}{c|}{\textbf{3D}} & 
\multicolumn{2}{c|}{\textbf{Albedo Front}} & 
\multicolumn{2}{c|}{\textbf{Albedo Back}} &
\textbf{Normals Front} &
\textbf{Normals Back} &
\textbf{Shaded Front}\\
&
Type & View &
CD $\downarrow$  & NC $\uparrow$ &
LPIPS $\downarrow$ & PSNR $\uparrow$ &
LPIPS $\downarrow$ & PSNR $\uparrow$ & 
Ang.\ $\downarrow$ & 
Ang.\ $\downarrow$ & 
PSNR $\uparrow$ \\
\hline
\hline
\multirow{4}{0.15\columnwidth}{Best of $N = 5$}
& $\mathbf{A}, \mathbf{N}$ & $F, B$ &
\csecond{1.11} & 0.84 & 0.14 & 21.39 & \csecond{0.26} & 18.74 & 20.24 & 23.86 & 25.13 \\
& $\mathbf{A}, \mathbf{D}$ & $F, B$ &
\cbest{1.05} & 0.78 & 0.14 & 21.68 & 0.27 & \csecond{19.57} & 19.14 & \cbest{22.39} & 24.94 \\
& $\mathbf{A}, \mathbf{N}, \mathbf{D}$ & $F$ &
1.18 & \csecond{0.86} & \csecond{0.12} & \csecond{23.04} & 0.27 & 19.17 & \csecond{19.13} & 22.59 & \csecond{25.97} \\
& $\mathbf{A}, \mathbf{N}, \mathbf{D}$ & $F, B$ &
1.16 & \cbest{0.86} & \cbest{0.11} & \cbest{23.26} & \cbest{0.22} & \cbest{21.05} & \cbest{19.12} & \csecond{22.55} & \cbest{27.09}\\
\hline
\multirow{4}{0.15\columnwidth}{Mean $\pm$ Std. $N = 5$}
& $\mathbf{A}, \mathbf{N}$ & $F, B$ &
\csecond{1.31} $\pm$ 0.8 & 0.82 $\pm$ 0.06 & 0.15 $\pm$ 0.04 & 20.96 $\pm$ 2.22 & 0.27 $\pm$ 0.08 & 18.31 $\pm$ 2.93 & 21.69 $\pm$ 4.34 & 24.65 $\pm$ 5.63 & 24.93 $\pm$ 2.07\\
& $\mathbf{A}, \mathbf{D}$ & $F, B$ &
\cbest{1.29} $\pm$ 1.0 & 0.76 $\pm$ 0.10 & 0.15 $\pm$ 0.04 & 21.07 $\pm$ 2.33 &  0.28 $\pm$ 0.08 & \csecond{18.83} $\pm$ 2.70 & 20.25 $\pm$ 3.62 & \cbest{22.89} $\pm$ 6.17 & 24.46 $\pm$ 2.68 \\
& $\mathbf{A}, \mathbf{N}, \mathbf{D}$ & $F$ &
1.37 $\pm$ 0.8 & \csecond{0.83} $\pm$ 0.06 & \csecond{0.13} $\pm$ 0.03 & \csecond{21.67} $\pm$ 2.30 & 0.28 $\pm$ 0.09 & 18.70 $\pm$ 2.71 & \cbest{19.75} $\pm$ 3.85 & 23.27 $\pm$ 6.21 & \csecond{25.88} $\pm$ 2.13\\
& $\mathbf{A}, \mathbf{N}, \mathbf{D}$ & $F, B$ &
1.38 $\pm$ 0.9 & \cbest{0.84} $\pm$ 0.05 & \cbest{0.12} $\pm$ 0.04 & \cbest{22.65} $\pm$ 2.43 & \cbest{0.24} $\pm$ 0.08 & \cbest{20.46} $\pm$ 3.37 & \csecond{20.24} $\pm$ 3.14 & \csecond{23.20} $\pm$ 5.54 & \cbest{26.82} $\pm$ 1.90

\end{tabular}
}
\caption{
Quantitative comparison between different types observation sets $\boldsymbol x_0$ used during implicit surface diffusion. $\mathbf{A}$, $\mathbf{N}$ and $\mathbf{D}$ refer to albedo, surface normal and depth images respectively. $F$ and $B$ designate front and back views. Populating $\boldsymbol x_0$ with front and back views of all 3 observation types gives the best all-round performance. Thus, this is the protocol used in the default DiffHuman model presented in the main paper. The \best{best} and \second{second best} results are marked. We report both best-of-$N$ and mean ($\pm$ std.) metrics for completeness.
}

\label{tab:supmat_obs}
\end{table*}

\section{Ablation Studies}
\label{sec:supmat_ablation}

This section presents additional ablation studies. We begin with a discussion of ``best-of-$N$'' vs.\ mean metrics, and report means and standard deviations to complement the best-of-$N$ metrics given in the main paper. Then, we provide a more detailed comparison of diffusion-via-rendering vs.\ our novel hybrid diffusion framework. We investigate different types of observation sets for diffusion, by dropping particular observations from $\boldsymbol x_0$. Finally, we implement classifier-free diffusion guidance \cite{ho2022ClassifierFreeDG} with an unconditional model and report corresponding metrics. 

\nbf{Mean vs.\ best-of-$N$ metrics}
The main paper reports evaluation metrics using the best-of-$N \in \{1, 5, 10\}$ reconstructions. This is justified for ambiguous metrics that measure performance in ill-posed tasks, where the ground-truth is \emph{but one} plausible solution. Our method is able to yield other solutions that are consistent with the input image but differ from the ground truth. Capturing the ground-truth within the range of solutions modelled by our predicted distributions is sufficient -- this is measured by best-of-$N$. 

However, not all metrics correspond to ill-posed tasks. In particular, ``shaded front'' metrics (\eg PSNR) measure the match between 3D reconstruction samples and the input image. \emph{All} samples should be input-consistent; hence, reporting the mean over $N$ samples is logical. This is arguably also true for LPIPS, which measures perceptual similarity, as noted by \cite{szymanowicz23viewset_diffusion}. Therefore, \cref{tab:supmat_gen_vs_rend,tab:supmat_obs} in this supplementary material report means and standard deviations, in addition to best-of-$N$ metrics. For completeness, these are provided for both well-posed and ill-posed tasks. Note that standard deviations are generally higher for back albedo and normals than the front. This is desired, and signifies greater diversity in unseen regions. Furthermore, standard deviations are lower for front shading, which should consistently match the conditioning image.

\nbf{Hybrid implicit surface diffusion}
\cref{tab:ablation_gen_vs_rend} in the main paper gives a brief ablation of our novel hybrid implicit surface diffusion framework. We provide more detailed results in \cref{tab:supmat_gen_vs_rend} in this supplementary material, where we compare denoising via $\mathtt{render}$, via $\mathtt{generate}$, and using a combination of both. The performances of all these methods are comparable, suggesting that the $\mathtt{generate}$ neural network has learned to imitate explicit rendering well. However, $\mathtt{generate}$ has a much reduced runtime -- specifically giving a $55\times$ speed-up over the reverse process. A qualitative comparison of these denoising strategies is visualised in \cref{fig:supmat_rend_vs_gen} in this supplementary material.

\nbf{Observations in $\boldsymbol x_0$} DiffHuman models a distribution over image-based, pixel-aligned observations of an implicit 3D surface $\mathcal{S}$. The default method utilises three types of observations of the front and back surfaces of $\mathcal{S}$: (i) unshaded albedo colour images $\mathbf{A}^F$ and $\mathbf{A}^B$, (ii) surface normal images $\mathbf{N}^F$ and $\mathbf{N}^B$ and (iii) depth maps  $\mathbf{D}^F$ and $\mathbf{D}^B$. In this supplementary material, we investigate the importance of each of these observations. Specifically, we train 3 ablation models by omitting depth, normals and back views in turn from the observation set $\boldsymbol x_0$. A quantitative comparison is provided in \cref{tab:supmat_obs}. Utilising all of aforementioned observation types in $\boldsymbol x_0$ gives the best all-round performance on a range of metrics. Dropping back views intuitively worsens metrics computed with back renders. Omitting depth and normals also generally degrades performance - apart from Chamfer distance. However, we note that Chamfer distance is a noisy metric, as evidenced by the large relative standard devitions, and it is difficult to make conclusive judgements from these results.

\paragraph{Classifier-free guidance} \cite{ho2022ClassifierFreeDG} is an inference-time technique used to trade-off sample quality (including input-consistency) vs.\ diversity in conditional diffusion models. We experimented with applying guidance to our method, by jointly training a conditional and an unconditional implicit surface diffusion model. In practice, this was achieved by randomly dropping the conditioning image $\mathbf{I}$ as a network input with probability $0.2$. Quantitative results are reported in  \cref{tab:supmat_cfg} in this supplementary material. We found that guidance with an unconditional model can indeed improve the match between 3D reconstruction samples and the conditioning image, as measured by metrics corresponding to shaded front renders. However, it caused a deterioration of most other metrics -- shown by row 1 vs.\ row 2 in  \cref{tab:supmat_cfg}. Moreover, training with random condition dropping yielded worse performance than a model that always sees a conditioning image. Perhaps a larger and more diverse training dataset is needed to fully realise the benefits of diffusion guidance in this task. Nevertheless, we find it instructive to visualise unconditional generation samples in \cref{fig:supmat_uncond} in this supplementary material. These exhibit a significant amount of diversity, covering a range of clothing and hair styles, colours and geometries. Moreover, unconditional samples are generated starting from random noise in the silhouette of a particular body shape (see \cref{fig:supmat_uncond}). This is a by-product of the fact that our method applies foreground masking to all neural network inputs. Noise within a silhouette can be considered as a form of implicit conditioning, and allows us to exert control over the body shapes of 3D human samples.

\begin{table*}[t]
\centering
\resizebox{\linewidth}{!}{
\begin{tabular}{ l | l | c c | c c | c c | c | c | c}
\multirow{2}{0.25\columnwidth}{\textbf{Train with Drop Cond.}} &
\multirow{2}{0.2\columnwidth}{\textbf{Test with Guidance}} &
\multicolumn{2}{c|}{\textbf{3D}} & 
\multicolumn{2}{c|}{\textbf{Albedo Front}} & 
\multicolumn{2}{c|}{\textbf{Albedo Back}} &
\textbf{Normals Front} &
\textbf{Normals Back} &
\textbf{Shaded Front}\\
&
&
CD $\downarrow$  & NC $\uparrow$ &
LPIPS $\downarrow$ & PSNR $\uparrow$ &
LPIPS $\downarrow$ & PSNR $\uparrow$ & 
Ang.\ $\downarrow$ &
Ang.\ $\downarrow$ & 
PSNR $\uparrow$ \\
\hline
\hline
Yes & No &
\csecond{1.29} & \csecond{0.84} & \csecond{0.13} & \csecond{22.22} & \csecond{0.25} & \csecond{20.37} & \csecond{20.38} & \csecond{22.90} & 26.07 \\
Yes & Yes &
1.42 & 0.82 & 0.14 & 21.37 & \csecond{0.25} & 19.98 & 22.25 & 24.31 &  \csecond{26.99} \\
No & No &
\cbest{1.16} & \cbest{0.86} & \cbest{0.11} & \cbest{23.26} & \cbest{0.23} & \cbest{21.06} & \cbest{19.11} & \cbest{22.46} & \cbest{27.09} \\

\end{tabular}
}
\caption{
Quantitative evaluation of classifier-free diffusion guidance \cite{ho2022ClassifierFreeDG} applied to DiffHuman. We jointly train an unconditional and conditional implicit surface diffusion model, by randomly dropping the conditioning image $\mathbf{I}$. The effectiveness of guidance with the unconditional model is evaluated in rows 1 and 2. Guidance improves the match between 3D reconstruction samples and the conditioning image, as measured by ``Shaded Front'' metrics. However, it causes a deterioration of most other metrics. In addition, we report results from the standard DiffHuman model trained without random condition dropping in row 3. This consistently outperforms the model trained with dropping. All metrics are best-of-$N=5$. We use a guidance weight of $3$. The \best{best} and \second{second best} results are marked.
}

\label{tab:supmat_cfg}
\end{table*}

\textbf{%
\begin{figure*}[tp]
    \centering
    \includegraphics[width=\linewidth,trim={0 0.3cm 0 0},clip]{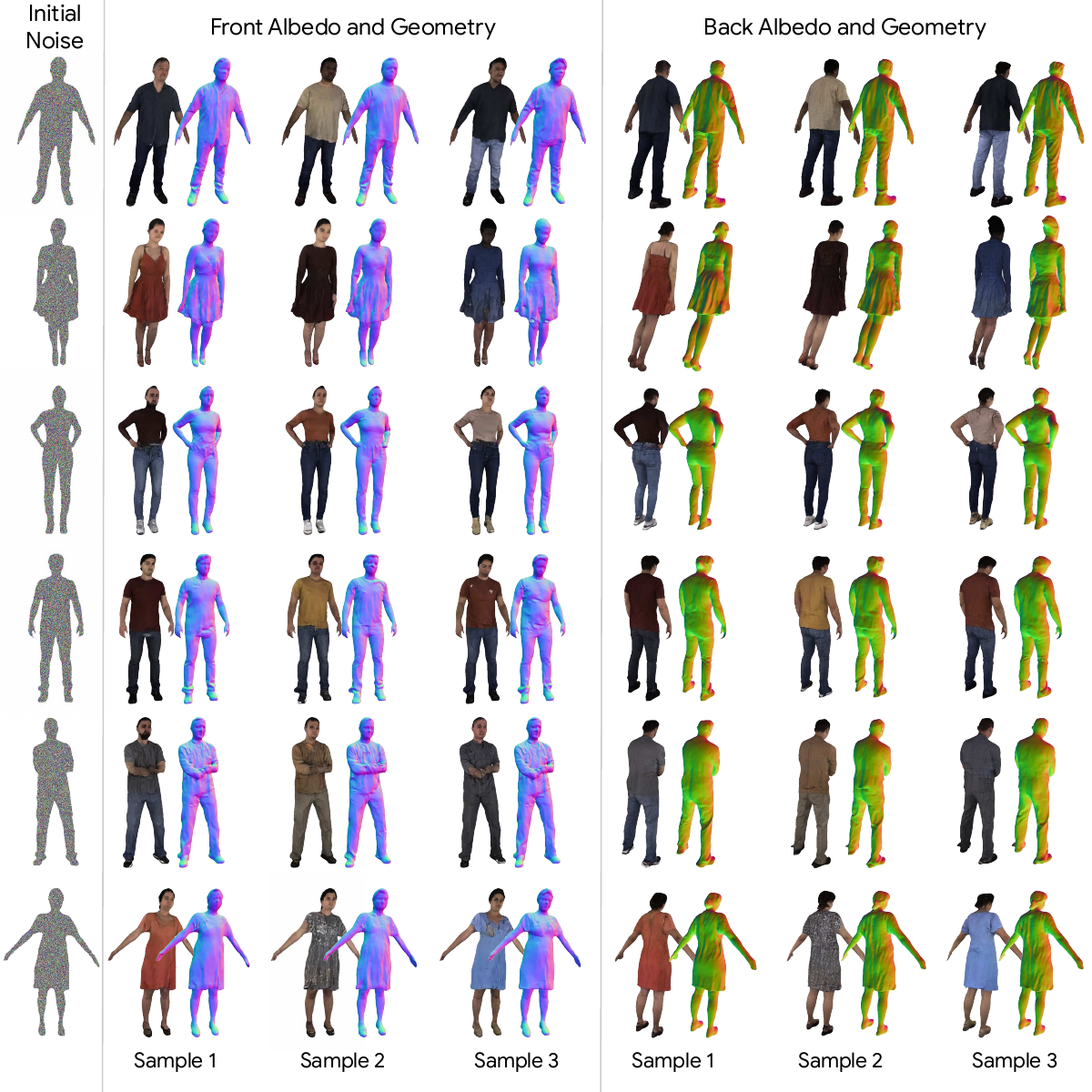}
    \caption{\textbf{Unconditional 3D reconstruction samples} generated using an implicit surface diffusion model trained with random condition dropping (following the protocol of classifier-free guidance \cite{ho2022ClassifierFreeDG}). These unconditional samples are generated from random noise only, which is masked using a silhouette in the shape of the desired subject. They exhibit significant diversity in terms of clothing styles, colours and geometries, as well as hairstyles, facial features and skin tones. For more ambiguous body shapes, different gendered properties are visible. The silhouette masking can be considered as a form of implicit conditioning, and allows us to exert some control over the 3D samples. Faces and certain body parts are blurrier for these unconditional samples than the conditional samples visualised in other figures. This is somewhat unsurprising, since conditioning images carry a lot of information on these fine features, which unconditional samples are not privy to.}
    \label{fig:supmat_uncond}
\end{figure*}
}

\textbf{%
\begin{figure*}[tp]
    \centering
    \includegraphics[width=\linewidth,trim={0 2.2cm 0 0},clip]{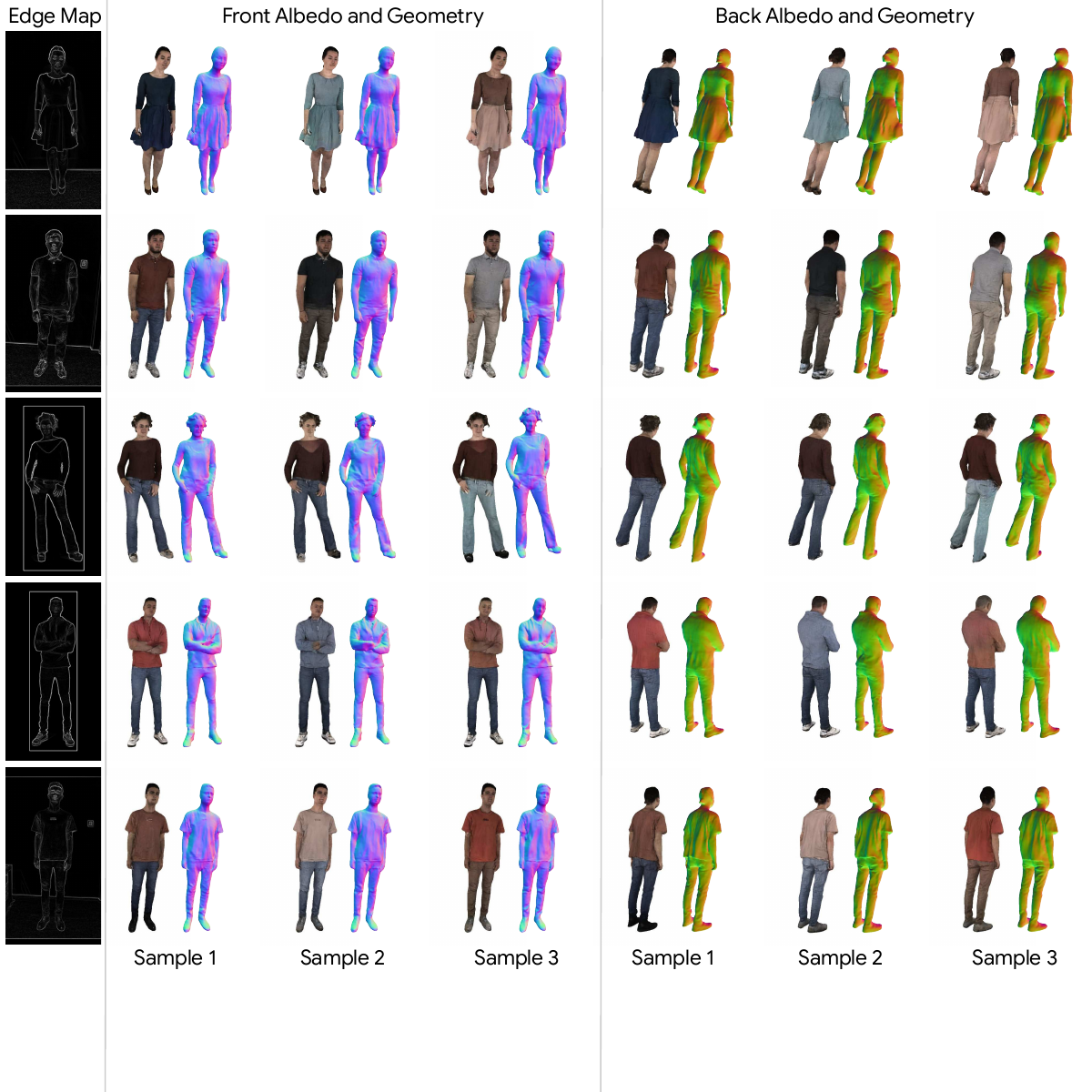}
    \caption{
    \textbf{3D reconstruction samples conditioned on edge map inputs.} These samples are generated using an implicit surface diffusion model that was pre-trained with conditioning RGB images, and then fine-tuned using conditioning edge maps -- inspired by ControlNet \cite{zhang2023controlnet}. Edges are obtained as image gradients using the Sobel operator. The 3D samples exhibit diverse colours, while the surface geometry respects the edge maps. This experiment demonstrates that samples from DiffHuman can be controlled via simpler conditioning inputs than full RGB images, which opens the possibility for generative applications beyond reconstruction from monocular images.
    }
    \label{fig:supmat_edge_cond}
\end{figure*}
}

\textbf{%
\begin{figure*}[tp]
    \centering
    \includegraphics[width=\linewidth,trim={0 1.7cm 0 0},clip]{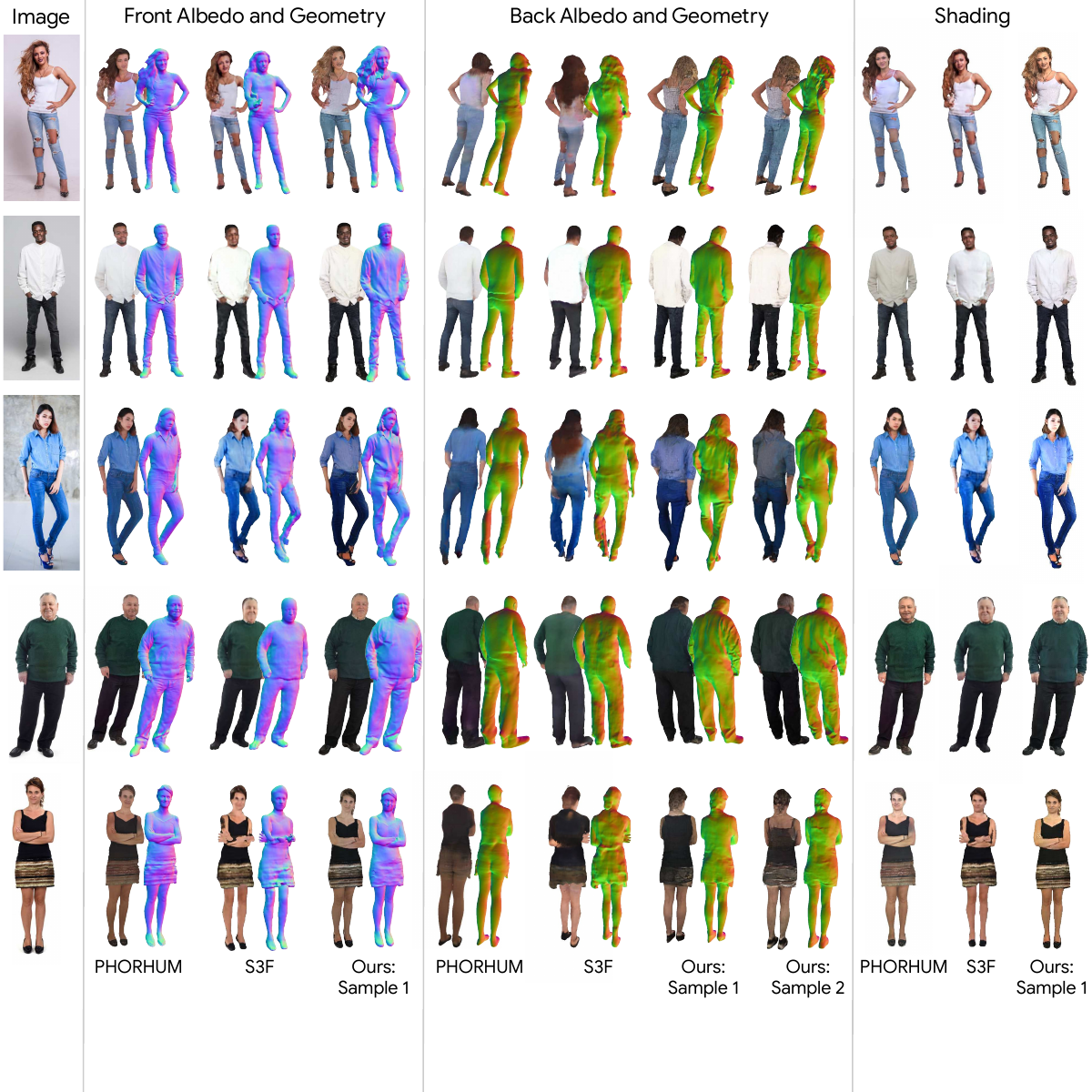}
    \caption{\textbf{Qualitative comparison against deterministic monocular 3D human reconstruction methods that predict geometry, surface albedo and shaded colour: PHORHUM \cite{alldieck2022phorhum} and S3F \cite{corona2023s3f}.} We show results from the original PHORHUM paper -- not our retrained version. Our method, DiffHuman, predicts a distribution over 3D reconstructions from which we can draw multiple samples. We visualise 2 samples from the back and 1 sample from the front for our method. PHORHUM outputs good front predictions, but exhibits flat geometry and blurry colours on the back. S3F \cite{corona2023s3f} yields more detailed geometry, but colours are still often blurry. Moreover, shaded renders of the reconstructions from each of these methods do not consistently match the input image. Our method is able to output multiple samples that are detailed, both in seen and unseen regions. In particular, note the hair geometry in row 1 and  diversity of dress styles (from the back) in row 5. Samples from our method exhibit a greater level of input-consistency, as shown by the shaded renders in rows 1, 2 and 4. Furthermore, we can faithfully handle a wider variety of body shapes, such as row 4.}
    \label{fig:supmat_compare_color}
\end{figure*}
}

\textbf{%
\begin{figure*}[tp]
    \centering
    \includegraphics[width=\linewidth,trim={0 0.3cm 0 0},clip]{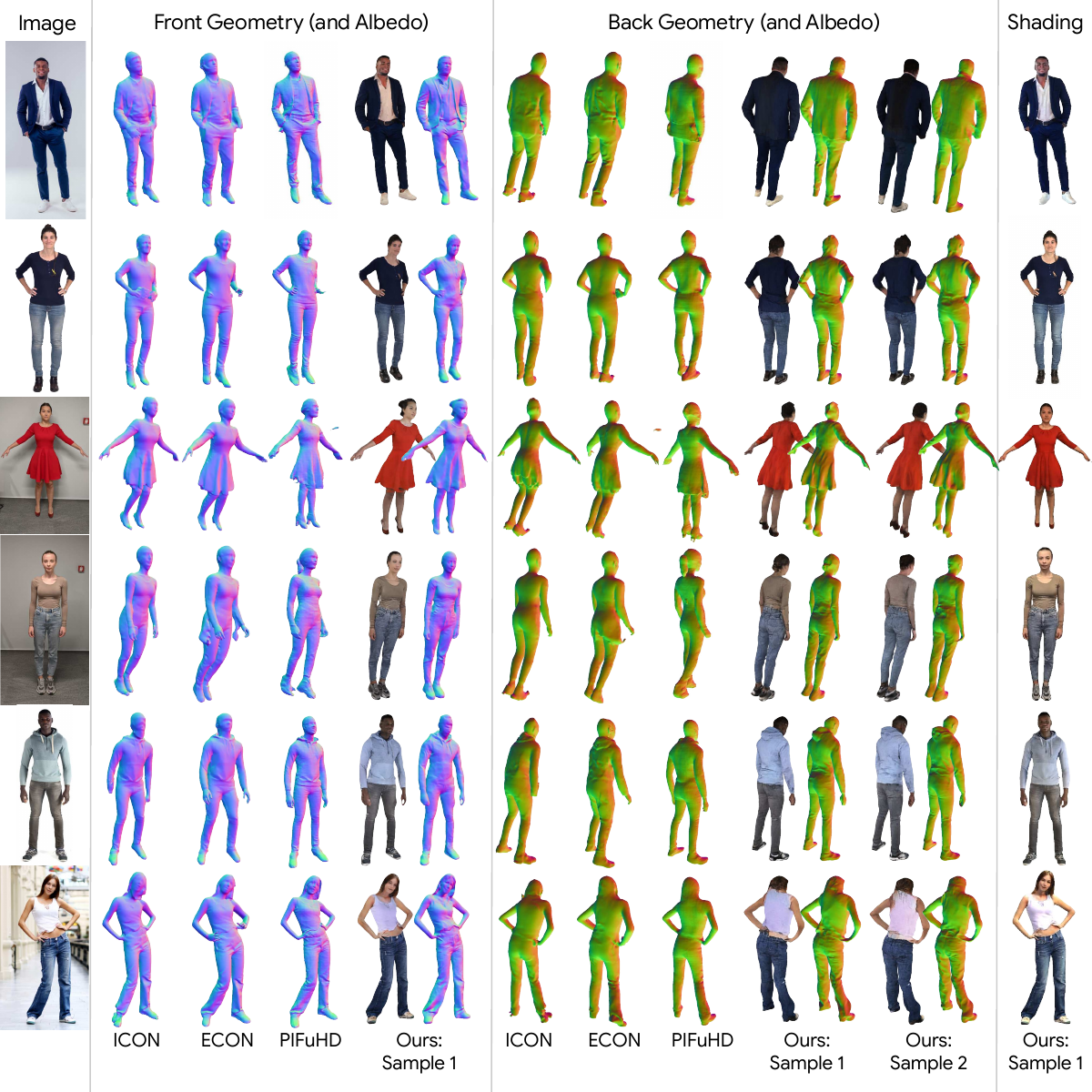}
    \caption{\textbf{Qualitative comparison against deterministic monocular 3D human reconstruction methods that predict only surface geometry: PIFuHD \cite{saito2020pifuhd}, ICON \cite{xiu2022icon} and ECON \cite{xiu2023econ}.} Our method, DiffHuman, predicts a distribution over 3D reconstructions from which we can draw multiple samples. We visualise 2 samples from the back and 1 sample from the front for our method. Samples from our method exhibiti greater geometric detail, both in seen and unseen and regions. In particular, note the front of the suit jacket in row 1, skirt in row 3, trousers in row 4 and hood in row 5. Moreover, when such details are unlikely -- \eg the back of the jacket in row 1, which is typically flat -- our method plausibly outputs samples with simpler geometry. Samples differ in hair styles and clothing colours on the back.}
    \label{fig:supmat_compare_nocolor}
\end{figure*}
}

\textbf{%
\begin{figure*}[tp]
    \centering
    \includegraphics[width=0.85\linewidth,trim={0 3.8cm 0 0},clip]{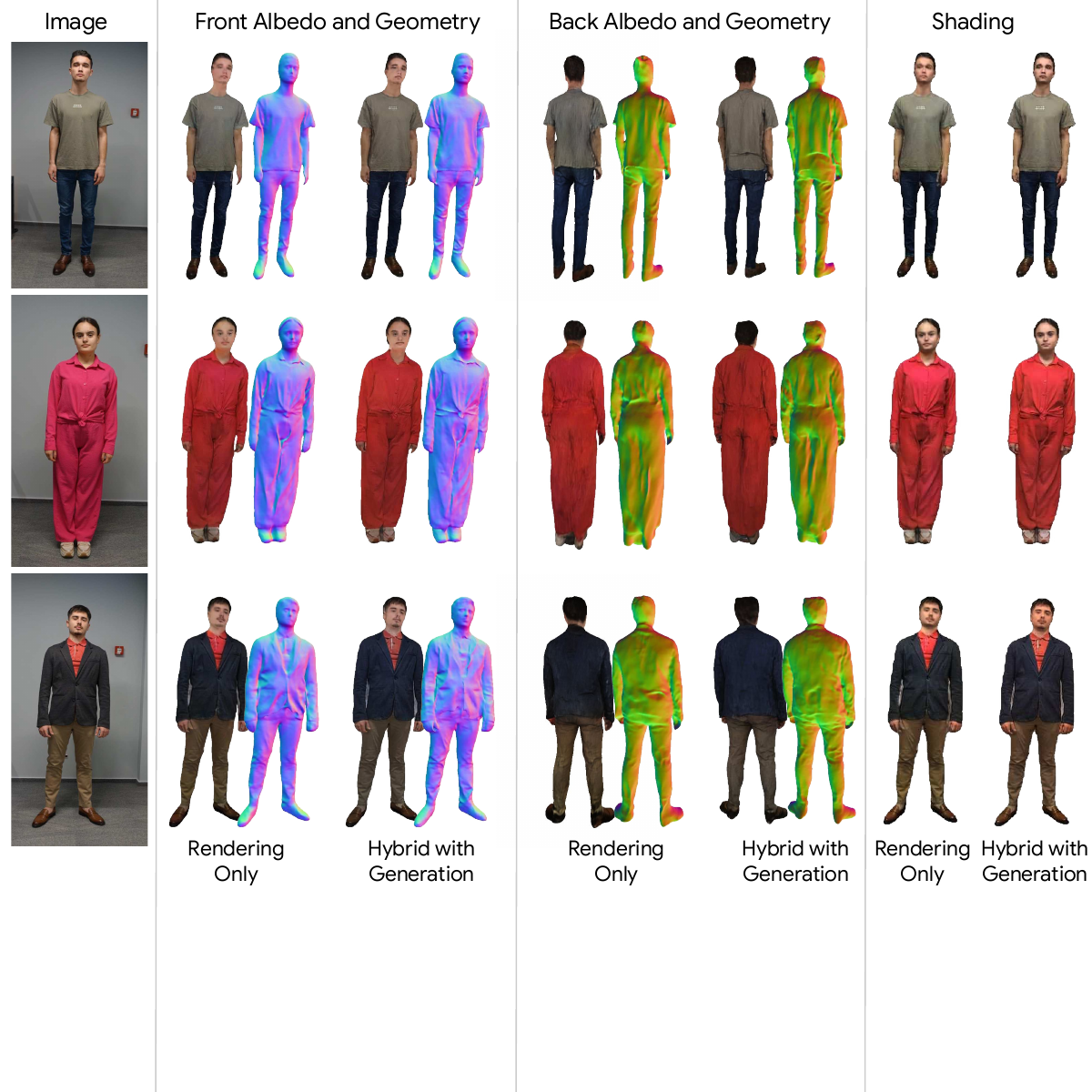}
    \caption{\textbf{Qualitative comparison between implicit surface diffusion via rendering, and hybrid diffusion using both rendering and generation.} Diffusion via rendering involves rendering an intermediate 3D representation in each denoising diffusion step to obtain a denoised sample. Hybrid diffusion uses a generator network that imitates rendering during the denoising process, at a much faster runtime. This figure complements Table \ref{tab:ablation_gen_vs_rend} in the main manuscript and Table \ref{tab:supmat_gen_vs_rend} in this supplement, by showing that samples from both these denoising processes are similar -- quantitatively and qualitatively. This suggests that the generator network learns to imitate explicit rendering sufficiently well. In fact, samples obtained via generation are often perceptually preferable to rendered samples (see the face in row 2). This could be because the generator network focuses solely on synthesising realistic observations, and is not constrained by explicit 3D geometry.}
    \label{fig:supmat_rend_vs_gen}
\end{figure*}
}

\newpage
\section{Qualitative Results}
\label{sec:supmat_quali}

This section provides further qualitative comparisons with current deterministic approaches to photorealistic 3D human reconstruction. \cref{fig:supmat_compare_color} visualises samples from DiffHuman against  reconstructions from PHORHUM \cite{alldieck2022phorhum} and S3F \cite{corona2023s3f} -- both of which estimate surface geometry, albedo colour and illumination-dependent shading.  \cref{fig:supmat_compare_nocolor} compares DiffHuman with methods that only estimate surface geometry: PIFuHD \cite{saito2020pifuhd}, ICON \cite{xiu2022icon} and ECON \cite{xiu2023econ}. 

Furthermore, we present qualitative results from additional experiments investigating the feasibility of DiffHuman as a generative model. As mentioned previously, \cref{fig:supmat_uncond} visualises unconditional 3D human samples generated from random noise in the silhouette of a given body shape. This allows us to loosely control the shape of 3D human samples. We extend this approach, by experimenting with using edge maps as conditioning images -- inspired by ControlNet \cite{zhang2023controlnet}. This allows us to have more fine-grained control over 3D samples, without having to provide a full RGB conditioning image. Qualitative results are given in \cref{fig:supmat_edge_cond}. These serve as a proof-of-concept for controllable generative applications beyond reconstruction.

Finally, \cref{fig:supmat_rend_vs_gen} compares implicit surface diffusion via $\mathtt{render}$ vs.\ $\mathtt{generate}$, to support the ablation studies presented in \cref{tab:supmat_gen_vs_rend}.

\fi

\newpage
\twocolumn[]
{\small
\bibliographystyle{ieeenat_fullname}
\bibliography{11_references}
}

\end{document}